\pdfoutput=1
\documentclass[11pt]{style/prism-princeton}

\usepackage[utf8]{inputenc}
\usepackage{url}
\usepackage{nicefrac}
\usepackage{array}
\usepackage{makecell}
\usepackage{dsfont}
\usepackage{subcaption}
\usepackage{tablefootnote}
\usepackage{algorithm}
\usepackage{wrapfig}

\newcommand{\ours}{\textsc{RAFT}}

\newcolumntype{C}[1]{>{\centering\arraybackslash}p{#1}}

\makeatletter
\renewenvironment{figure*}[1][]{\@dblfloat{figure}[!tbp]}{\end@dblfloat}
\renewenvironment{table*}[1][]{\@dblfloat{table}[!tbp]}{\end@dblfloat}
\renewenvironment{figure}[1][]{\@float{figure}[!tbp]}{\end@float}
\AtBeginDocument{\@ifpackageloaded{hyperref}{\hypersetup{pdftitle={RAFT: Reasoning-Aligned Fine-Tuning Improves Robustness of VLA Post-Training}, pdfauthor={Shresth Grover, Andrey Kolobov, Dhruv Shah}}}{}}
\makeatother

\author[1]{Shresth Grover}
\author[2]{Andrey Kolobov}
\author[1]{Dhruv Shah}
\affiliation[1]{Princeton University}
\affiliation[2]{Microsoft Research}

\begin{document}

\title{RAFT: Reasoning-Aligned Fine-Tuning Improves Robustness of VLA Post-Training}

\abstract{
Adapting a generalist vision-language-action (VLA) policy to a target task with limited demonstrations can improve nominal task success while reducing robustness to variations in appearance, objects, and instructions. We introduce reasoning-aligned fine-tuning (RAFT), a two-stage post-training procedure: RAFT first trains low-rank adapters on task-relevant reasoning annotations, then freezes them while fine-tuning the policy on action demonstrations. On a physical DROID robot, RAFT achieves $75.0\%$ out-of-distribution success, compared with $27.9\%$ for plain fine-tuning; on a multi-stage bimanual YAM task, it reaches $70.0\%$ mean success, compared with $36.0\%$ for the strongest baseline. RAFT also improves success under visual, object, and instruction shifts in simulated pick-and-place and open-drawer tasks. Compared with plain fine-tuning, RAFT policies place more attention on annotated task regions and yield higher linear-probe accuracy for object identity in later layers. A broader reasoning-data mixture further improves downstream success across demonstration budgets.
}


\def\websitelist{}
\def\codelist{\vspace*{0.9cm}}
\website{https://princeton-prism.github.io/RAFT/}{princeton-prism.github.io/RAFT}
\code{https://github.com/princeton-prism/RAFT}{github.com/princeton-prism/RAFT}

\maketitle

\section{Introduction}

Generalist vision-language-action (VLA) policies, built on pretrained vision-language models and large robot datasets~\cite{anil2023palm,open_x_embodiment_rt_x_2023,khazatsky2024droid,kim2024openvla,pmlr-v305-black25a,bjorck2025gr00t}, typically still require adaptation to reach the success rates needed to deploy the policy on a real-world target task~\cite{black2024pi_0,kim2025fine}. Because collecting task-specific demonstrations is expensive, this post-training often relies on a small dataset collected in a narrow set of environments. Fine-tuning can improve nominal success while reducing robustness to visual and semantic variations absent from those demonstrations~\cite{gao2026taxonomy,zhang2025effective,wang2025generalization,xiang2025parallels,kaplanis2019policy,xing2025shortcut}. Existing approaches address this problem through parameter merging, as in RETAIN~\cite{yadav2025retain}, or regularization during adaptation~\cite{kachaev2025dontblind,huang2026breakinglockin}. We study a complementary approach, aligning a pretrained policy with task-relevant reasoning supervision before adapting it to the target actions.
\begin{figure}[t]
  \centering
  \includegraphics[width=\linewidth]{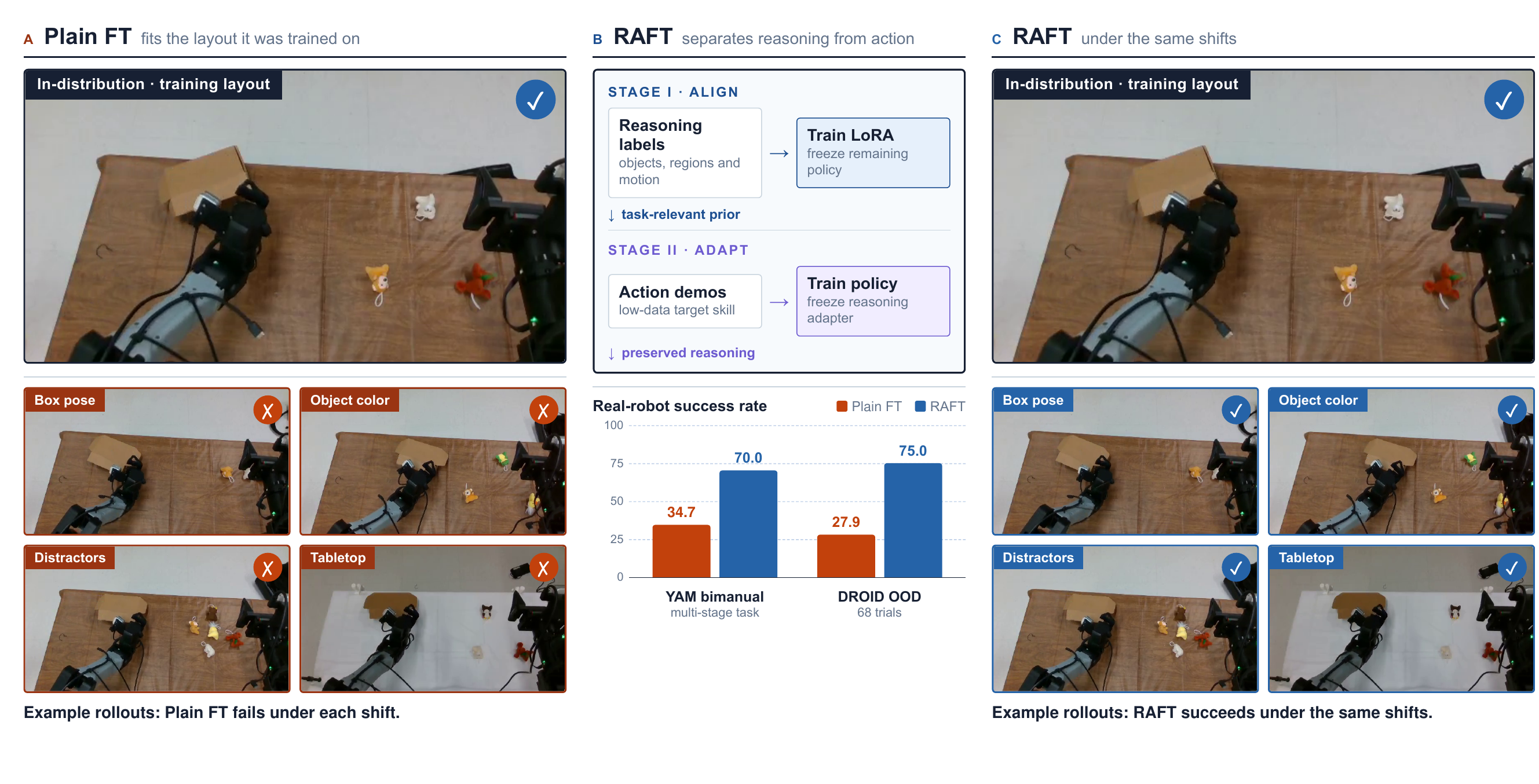}
  \caption{\textbf{Reasoning-aligned fine-tuning for robust VLA adaptation.} (a) Plain FT, fine-tuned only on actions, succeeds in the training layout but fails under small scene changes. (b) \ours{} first learns task-relevant reasoning through LoRA adapters, then freezes them while fine-tuning the policy on action demonstrations. (c) With the same demonstrations, \ours{} succeeds under these shifts. On physical robots, \ours{} raises success over Plain FT from $34.7\%$ to $70.0\%$ on YAM and from $27.9\%$ to $75.0\%$ on out-of-distribution DROID trials.}
  \label{fig:teaser}
\end{figure}

Reasoning supervision can teach policies to identify task-relevant objects, spatial relations, and motion structure~\cite{zawalski2024ecot,fang2026molmoact2,zha2026lap,li2025spatial}. We study how reasoning capabilities in pretrained VLMs can support low-data adaptation of generalist VLAs to a target skill. We introduce reasoning-aligned fine-tuning (\ours{}), a two-stage adaptation procedure. Stage~I learns low-rank adapters in the language and vision modules from task-relevant reasoning annotations while keeping pretrained parameters fixed. Stage~II freezes these adapters and fine-tunes the language model, action expert, and projection layers on action demonstrations. Action gradients therefore do not directly update the reasoning-adapter weights. We find that how reasoning supervision is incorporated matters: jointly co-training reasoning and actions underperforms plain fine-tuning ($63.2\%$ vs.\ $79.6\%$ mean success on simulated pick and place), and updating the reasoning adapter during action learning falls below Plain FT ($74.0\%$ vs.\ $79.6\%$) (Secs.~\ref{sec:cotrain-control} and~\ref{sec:stage-ablation}). Staging the two, with the adapter frozen, yields these robustness gains. \ours{} policies also place more attention on annotated task regions and yield higher linear-probe accuracy for object identity and episode-time position than plainly fine-tuned policies (Sec.~\ref{sec:diagnostics}).

We evaluate \ours{} on two simulated tasks and two physical platforms: the single-arm DROID setup~\cite{khazatsky2024droid} and the bimanual YAM setup~\cite{fang2026molmoact2}. Evaluations test visual, object, instruction, and geometric variations of the adapted tasks. \ours{} achieves the highest mean success on simulated pick and place and improves open-drawer success, with the largest gains on the physical robots: success on $75\%$ of out-of-distribution DROID trials versus $28\%$ for plain fine-tuning, and $70\%$ mean success on the multi-stage YAM task versus $36\%$ for the strongest baseline. Our contributions are threefold: (1) We propose RAFT a post training recipt that separates task relevant reasoning supervision from action fine tuning (2) We show improved downstream performace and robustness in simulation and on two physical robot platforms; and (3) We analyze why RAFT helps generalist policy adapt to a target by analyzing the features learned by RAFT.

\section{Related Work}

Generalist robot policies have made substantial progress in learning transferable skills and supporting deployment through post-training~\cite{bousmalis2023robocat,kim2024openvla,team2024octo,guo2025improving,li2025simplevla,zang2025rlinf,pan2026sop}. Within this setting, work on VLA fine-tuning studies reliable task execution~\cite{kim2025fine,zhang2025effective}, while FAST improves action tokenization for autoregressive VLAs~\cite{pertsch2025fast}. ControlVLA uses object-centric conditioning for few-shot adaptation~\cite{li2025controlvla}. These advances make specialization more effective, but nominal task success can still mask failures under distribution shift~\cite{gao2026taxonomy,zhou2025libero,fei2026libero}. This motivates studying robustness to variations absent from the target demonstrations.

Robust post-training methods seek to adapt generalist policies while retaining useful pretrained capabilities. Weight-space approaches interpolate or merge pretrained and adapted parameters~\cite{wortsman2022robust,jang2024model,dey2024revla}, while regularization and steerability-preserving adaptation constrain changes during fine-tuning~\cite{kachaev2025dontblind,grover2025enhancing,sharma2023lossless,huang2026breakinglockin}. RETAIN~\cite{yadav2025retain} interpolates pretrained and fine-tuned policy weights to improve generalization to variations of a new skill and retain broader capabilities. DeLock~\cite{huang2026breakinglockin} regularizes visual-encoder weight drift during post-training and uses contrastive prompt guidance at inference. \ours{} instead introduces task-relevant reasoning alignment before action fine-tuning, with complementary parameter freezing across the two stages.

Reasoning and grounding supervision provide a foundation for this approach. Embodied Chain-of-Thought interleaves sub-goal, spatial, and motion reasoning with action prediction~\cite{zawalski2024ecot}. Related work incorporates spatial representations, affordances, keypoints, and manipulable parts~\cite{fang2026molmoact2,zha2026lap,li2025spatial}, or object- and language-grounded supervision~\cite{pmlr-v305-black25a}. Steerable Policies use commands at multiple abstraction levels to control~\cite{chen2026steerable}, while VLS uses VLM-generated rewards to steer frozen policies at inference~\cite{liu2026vls}. Reasoning supervision is used to prepare that policy for specialization, and the learned adapters are held fixed during subsequent action learning. To the best of our knowledge, reasoning supervision has not been studied for low-data post-training of a pretrained generalist VLA. We evaluate this adaptation schedule through downstream robustness, additional training controls, and analyses of task-relevant representations.

\section{Method}

\begin{figure}[!tb]
  \centering
  \includegraphics[width=\textwidth,height=0.31\textheight,keepaspectratio]{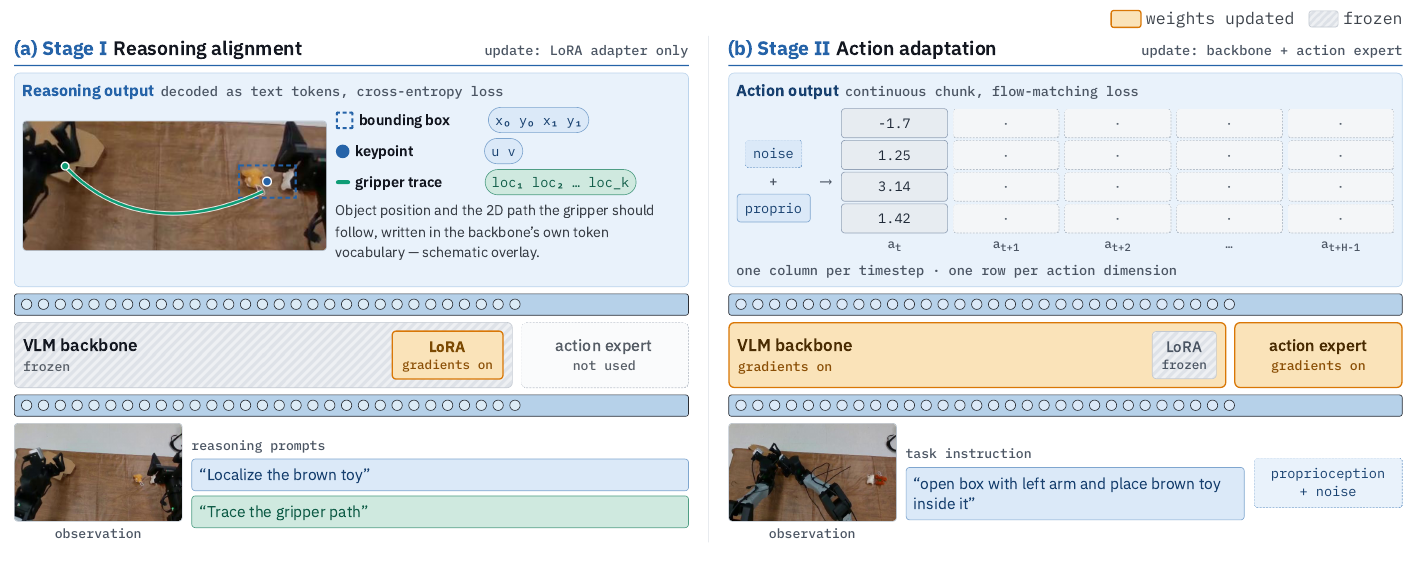}
  \caption{\textbf{Overview of \ours{}.} Stage~I updates only the reasoning adapters, using reasoning annotations. Stage~II freezes the adapters and updates the language model, action expert, and projection layers on action demonstrations, so action gradients do not reach the reasoning adapters.}
  \label{fig:method_overview}
\end{figure}

\subsection{Overview and Problem Setup}
We consider task adaptation from a pretrained generalist policy using a small demonstration dataset $\mathcal{D}_\eta=\{(s_t,a_t,T)\}$ collected in a narrow set of environments. Here $s_t$ contains the policy observations, $a_t$ is the demonstrated action, and $T$ is the task instruction.
We denote the Stage-II trainable non-adapter parameters (the language model, action expert, and projection layers) by $\theta$, initialized from $\theta_{\mathrm{pre}}$, and the LoRA parameters by $\phi$~\cite{hu2022lora}. In \ours{}, the pretrained SigLIP weights remain frozen throughout adaptation, while LoRA modules in both the vision encoder and language model are trained in Stage~I. Stage~II freezes these learned adapters and updates $\theta$. Thus, vision adaptation in \ours{} occurs through Stage-I LoRA rather than updates to the pretrained SigLIP weights. The resulting action policy is $\pi_{\theta,\phi}(a_t\mid s_t,T)$. Figure~\ref{fig:method_overview} summarizes the two stages; Sec.~\ref{app:training-details} gives the adapter placement and optimization settings.

\begin{figure*}[!tb]
\centering
\includegraphics[width=\linewidth]{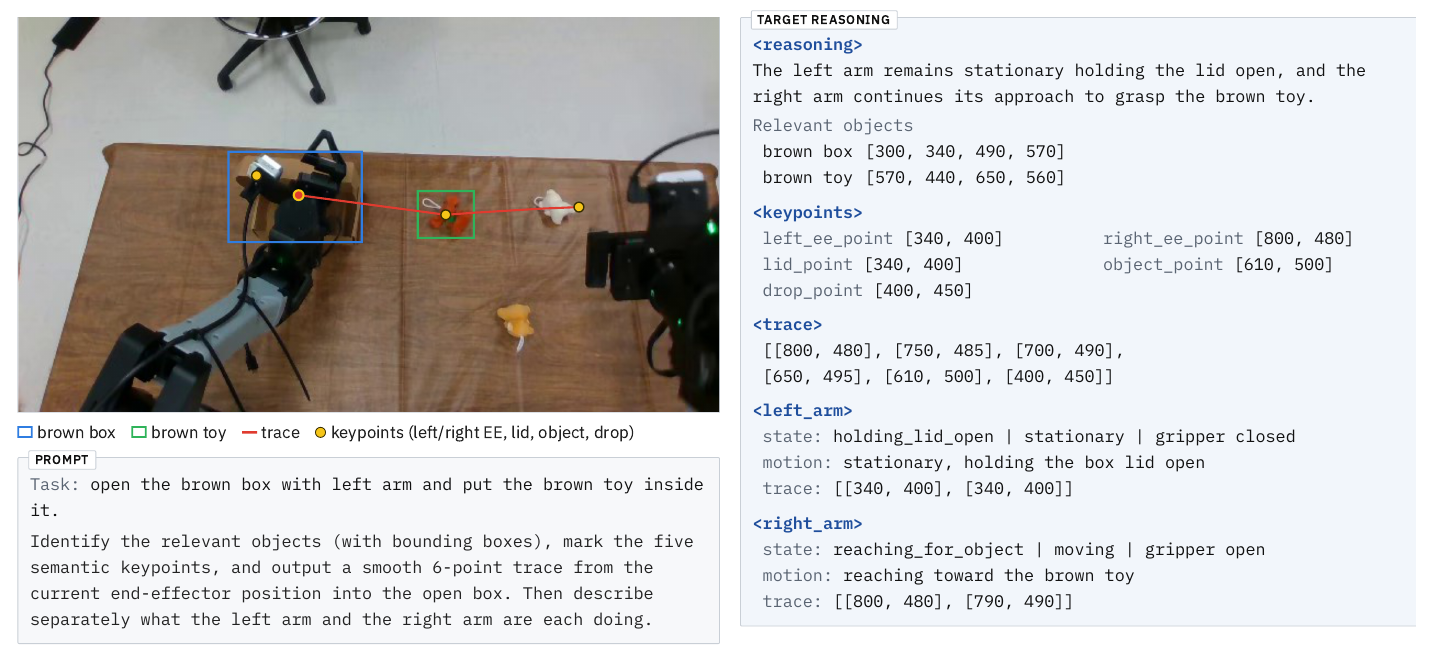}
\caption{\textbf{Example Stage-I reasoning annotation.} In this YAM example, the left arm holds the lid open while the right arm reaches for the toy. The target combines language, object boxes, keypoints, motion traces, and per-arm state information. Blue and green mark the box and toy, yellow marks keypoints, and red marks the active arm’s trace.}
\label{fig:reasoning_example}
\end{figure*}

\subsection{Stage I: Reasoning Alignment}
\label{sec:method-stage1}
Stage~I trains the adapter to predict task-relevant reasoning annotations. A VLM annotator (Gemini~3.1~Pro) produces task-relevant reasoning targets from observations and instructions on $\mathcal{D}_\eta$, yielding $\mathcal{D}_{\mathrm{reason}}=\{(s,r,T)\}$. These annotations describe the objects and interactions relevant to the task. The annotation choices include bounding boxes, keypoints, traces, spatial descriptions, and visual question answering. Section~\ref{subsec:modality} compares their effects. Each annotation is represented as a target token sequence $r=(r_1,\ldots,r_{|r|})$. We train the LoRA adapter using standard autoregressive cross-entropy:
\begin{equation}
\mathcal{L}_{\mathrm{reason}}(\phi)=
-\mathbb{E}_{(s,r,T)\sim\mathcal{D}_{\mathrm{reason}}}
\sum_{j=1}^{|r|}\log p_{\theta_{\mathrm{pre}},\phi}(r_j\mid s,T,r_{<j}).
\label{eq:reason}
\end{equation}
Only $\phi$ is updated in this stage; all remaining parameters are frozen. This supervision encourages task-relevant grounding before action adaptation.

\subsection{Stage II: Action Adaptation}
Stage~II turns the reasoning-aligned initialization into an action policy for the target task. We retain the learned adapter $\phi^\star$ and freeze its parameters. The Stage-II parameter set $\theta$ is then updated by supervised fine-tuning (SFT) on the task demonstrations~\cite{kim2024openvla,kim2025fine}:
\begin{equation}
\theta^\star=\arg\min_\theta\,
\underbrace{\mathbb{E}_{(s,a,T)\sim\mathcal{D}_\eta}
[\ell_{\mathrm{SFT}}(\pi_{\theta,\phi^\star};s,a,T)]}_{\mathcal{L}_{\mathrm{BC}}(\theta;\mathcal{D}_\eta,\phi^\star)}.
\label{eq:bc}
\end{equation}
Here $\ell_{\mathrm{SFT}}$ is the underlying policy's action-supervision objective. Stage~II updates the language model, action expert, and projection layers while the adapter and SigLIP vision tower remain frozen. At deployment, the resulting policy is $\pi_{\theta^\star,\phi^\star}$. The adapter ablation in the appendix (Sec.~\ref{sec:stage-ablation}) evaluates the benefit of freezing the reasoning adapter during action learning. Attention and linear-probe analyses (Sec.~\ref{sec:diagnostics}) provide empirical evidence of retained task grounding in the resulting policy.


\section{Experiments}
\label{sec:analysis}

The goal of our experiments is to evaluate \ours{}'s ability to retain task-relevant information and improve downstream robustness. Concretely, we aim to answer: (1) does \ours{} generalize better than baselines under distribution shift, in simulation and on physical robots? (2) do \ours{}'s representations retain more task-relevant information than a plainly fine-tuned policy, and does this track downstream success? (3) how does downstream performance scale with the amount of reasoning data used in Stage~I? and (4) How freezing Stage-I adapted compares with common approaches like co-training simulatenously?

\subsection{Experimental setup}
\label{subsec:setup}

\begin{figure*}[t]
\centering
\includegraphics[width=\textwidth]{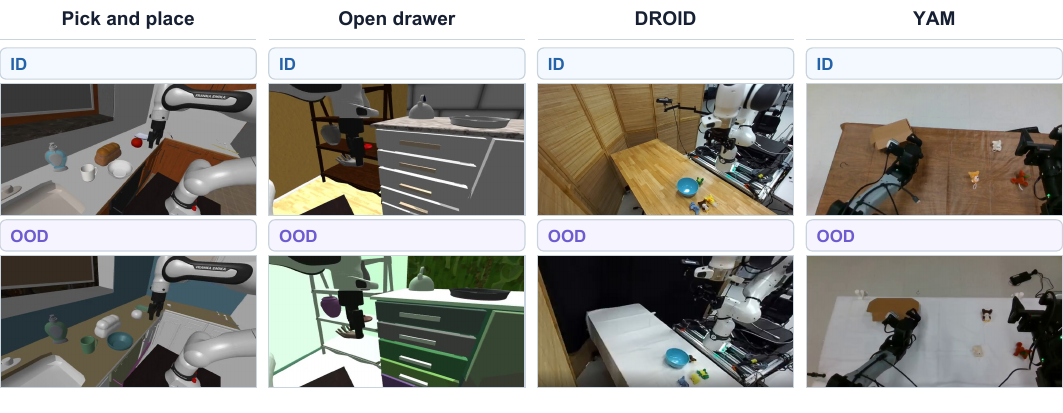}
\caption{\textbf{Evaluation scenes.} Representative in-distribution (ID) and out-of-distribution (OOD) observations, left to right: simulated pick and place, simulated open drawer, physical DROID pick and place, and the physical YAM two-stage task. The OOD panels show texture changes in simulation and background changes on the physical robots. The full set of evaluated shifts is shown in Fig.~\ref{fig:supp-evaluation-conditions}.}
\label{fig:evaluation-scenes}
\end{figure*}

\paragraph{Baselines}
We compare \ours{} with standard supervised fine-tuning (\emph{Plain~FT}), \emph{RETAIN}~\cite{yadav2025retain}, which merges pretrained and fine-tuned weights, and \emph{DeLock}~\cite{huang2026breakinglockin}, which regularizes visual-encoder weight drift during fine-tuning and applies contrastive prompt guidance at inference. We additionally report \emph{Base}, the pretrained generalist policy evaluated without post-training. All fine-tuned policies are evaluated after 5,000 action-training steps. Sec.~\ref{app:baseline-details} gives the baseline settings; Sec.~\ref{sec:cotrain-main} ablates the training design, with full adapter and co-training results in the appendix (Secs.~\ref{sec:stage-ablation} and~\ref{sec:cotrain-control}).

\begin{figure*}[!tb]
\centering
\begin{minipage}[t]{0.49\textwidth}
\vspace{0pt}\centering
\includegraphics[width=\linewidth]{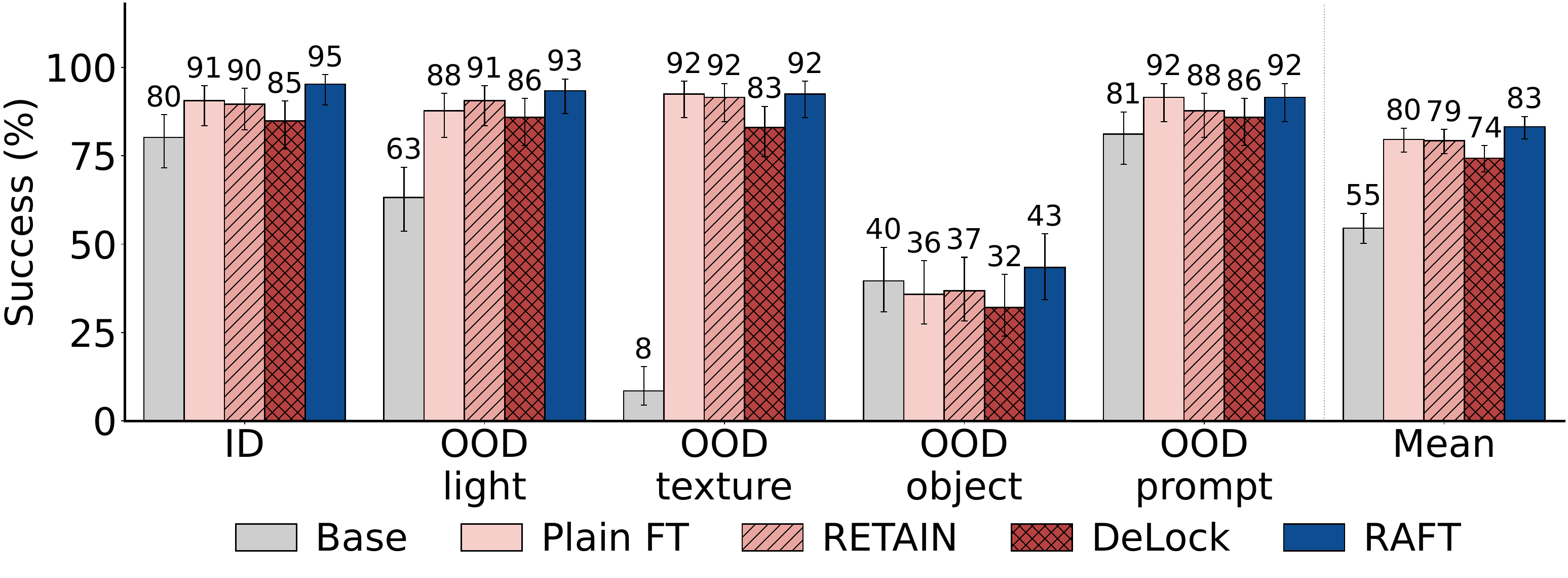}
\end{minipage}\hfill
\begin{minipage}[t]{0.49\textwidth}
\vspace{0pt}\centering
\includegraphics[width=\linewidth]{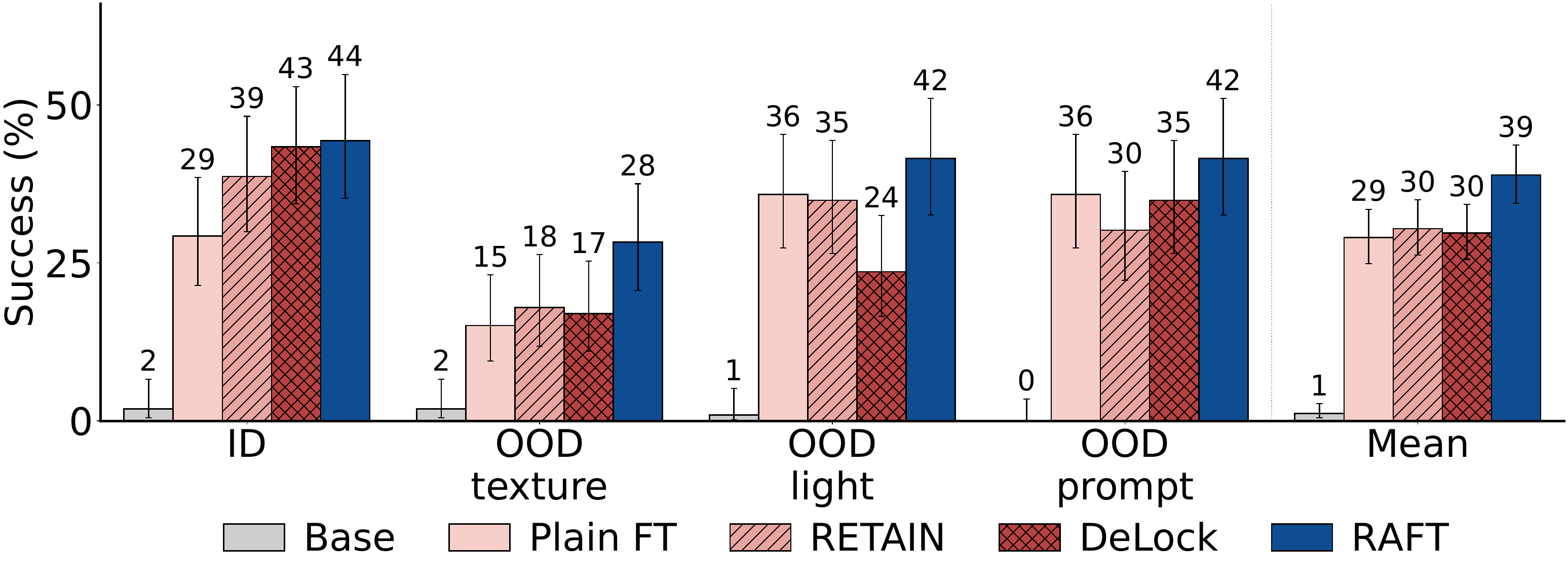}
\end{minipage}
\caption{\textbf{Simulation results under distribution shift.} Success rates for pick and place (left) and open drawer (right), with 106 trials per split. \ours{} uses full reasoning for pick and place and bounding-box supervision for open drawer (Sec.~\ref{subsec:modality}). \ours{} has the highest mean success on both tasks: $83\%$ versus $80\%$ for Plain FT on pick and place, and $39\%$ versus $29\%$ on open drawer.}
\label{fig:sim_results}
\end{figure*}

\subsection{\ours{} generalizes under distribution shift}
\label{sec:distribution-shift}

We evaluate \ours{} on pick-and-place and open-drawer tasks in simulation and on two real-world platforms: the single-arm DROID setup~\cite{khazatsky2024droid} and the dexterous, dual-arm YAM setup~\cite{fang2026molmoact2}, under the visual, object, and instruction or geometric shifts described below.

\paragraph{Simulation}
We evaluate MolmoBot~\cite{deshpande2026molmob0t} in the MolmoSpaces simulator~\cite{kim2026molmospaces} on pick-and-place and open-drawer tasks. For each task, we use 106 successful demonstrations from one layout provided in the MolmoSpaces benchmark for post-training. We measure robustness using perturbations absent from the demonstrations. \emph{OOD-light} changes the number, position, and intensity of light sources; \emph{OOD-texture} replaces table and scene textures; \emph{OOD-object} substitutes a semantically similar manipulation object unseen during post-training and is evaluated only for pick and place; and \emph{OOD-prompt} paraphrases the instruction without changing its meaning. \emph{ID} uses the demonstration layout. Each evaluation split contains $n=106$ trials with matched seeds across methods.

\paragraph{\ours{} has the highest simulated success, with the largest margin on novel objects}
On pick and place, \ours{} achieves $83\%$ mean success across the five evaluation splits, compared with $80\%$ for Plain FT, $79\%$ for RETAIN, and $74\%$ for DeLock (Fig.~\ref{fig:sim_results}). The largest margin is under object shift: $43\%$ for \ours{} versus $36\%$ for Plain FT and $37\%$ for RETAIN. On the open drawer task, \ours{} reaches $44\%$ success in ID and $28\%$ under texture shift, versus $29\%$ and $15\%$ for Plain FT, and it also leads under lighting and instruction shifts.

\begin{figure*}[!tb]
\centering
\begin{minipage}[t]{0.49\textwidth}
\vspace{0pt}\centering
\includegraphics[width=\linewidth]{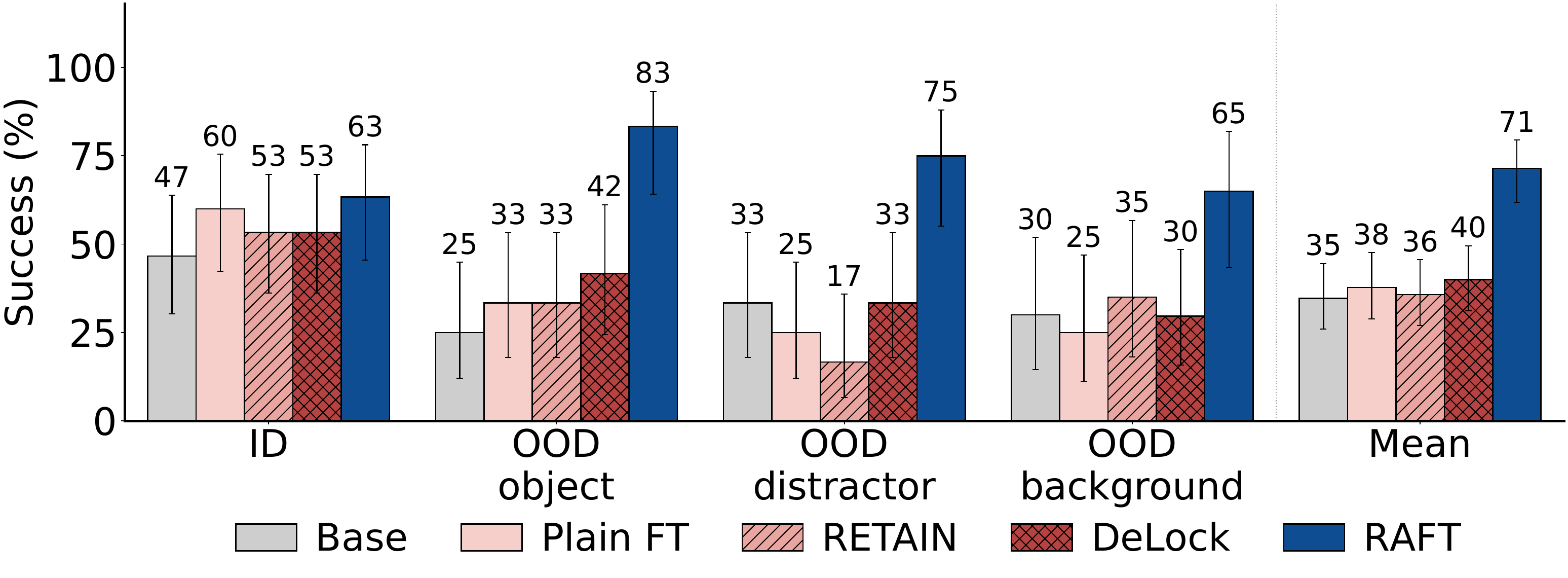}
\end{minipage}\hfill
\begin{minipage}[t]{0.49\textwidth}
\vspace{0pt}\centering
\includegraphics[width=\linewidth]{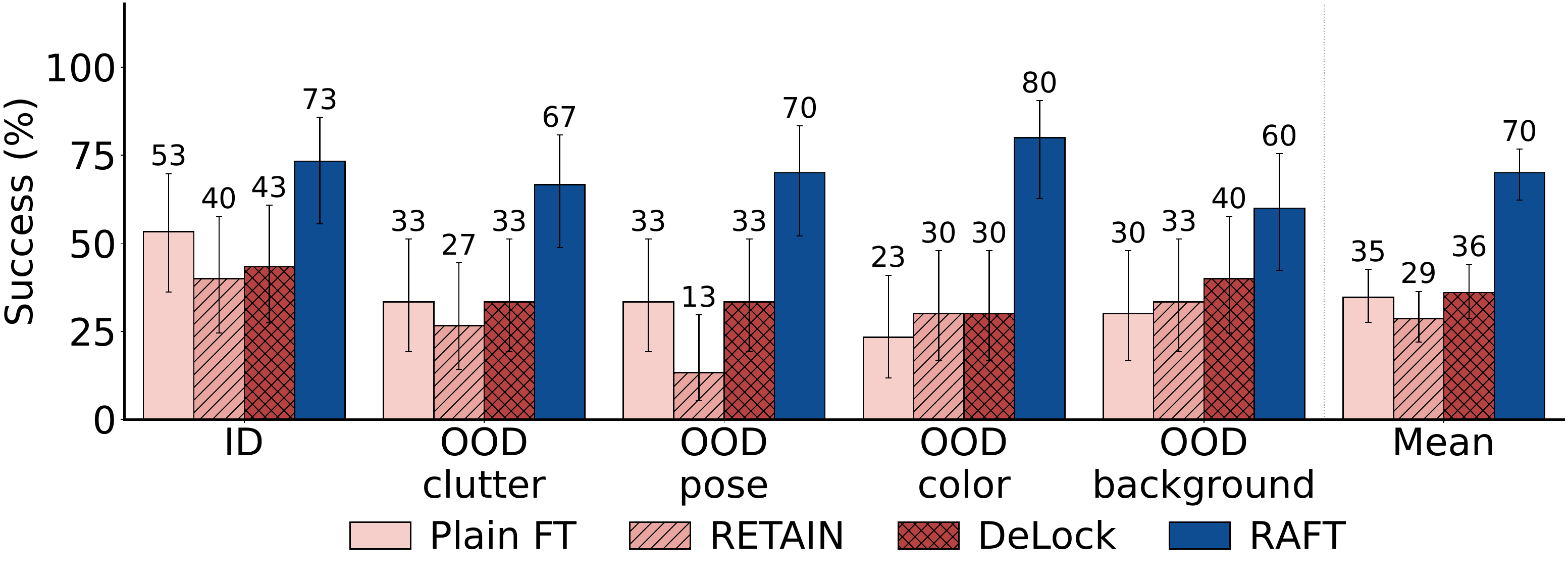}
\end{minipage}
\caption{\textbf{Real-robot results under distribution shift.} Left: DROID pick and place. Right: YAM box opening and place inside box. \ours{} has the highest success on both platforms: $75\%$ versus $28\%$ for Plain FT on out-of-distribution DROID trials, and $70\%$ versus $35\%$ on YAM.}
\label{fig:real_results}
\end{figure*}

\paragraph{Real-world: DROID}
We post-train MolmoBot~\cite{deshpande2026molmob0t} on 50 pick-and-place demonstrations and evaluate on a physical DROID robot under four axes: the demonstration setup (ID; blue, yellow, and brown toys), novel object colors (OOD-Object, e.g., a white toy in place of the brown one), added distractors (OOD-Distractor), and a changed tabletop surface (OOD-Background). On out-of-distribution DROID trials, \ours{} succeeds in $75.0\%$ of trials versus $27.9\%$ for Plain FT (Fig.~\ref{fig:real_results}): $83\%$ versus $33\%$ with novel object colors, $75\%$ versus $25\%$ with distractors, and $65\%$ versus $25\%$ on the changed tabletop. In the demonstration setup, the two are close ($63\%$ versus $60\%$).

\paragraph{Real-world: YAM}
For the dual-arm YAM platform, we post-train $\pi_{0.5}$~\cite{pmlr-v305-black25a} on 100 demonstrations of a long-horizon task: open a hinged box, pick up an object, and place it inside. Because placement depends on the box being open, the task separates completing the first interaction from completing the whole sequence. We evaluate in the demonstration setup and under four shifts: clutter, box pose, object color, and background.
On the bimanual YAM task, \ours{} completes the full sequence in $70.0\%$ of trials pooled over the five axes, versus $36\%$ for DeLock, $35\%$ for Plain FT, and $29\%$ for RETAIN (Fig.~\ref{fig:real_results}), and it has the highest success on every axis ($73\%$ in-distribution and $60\%$--$80\%$ under the four shifts). Plain FT usually opens the box but then fails to place the object. 

\begin{figure*}[t]
\centering
\includegraphics[width=\textwidth]{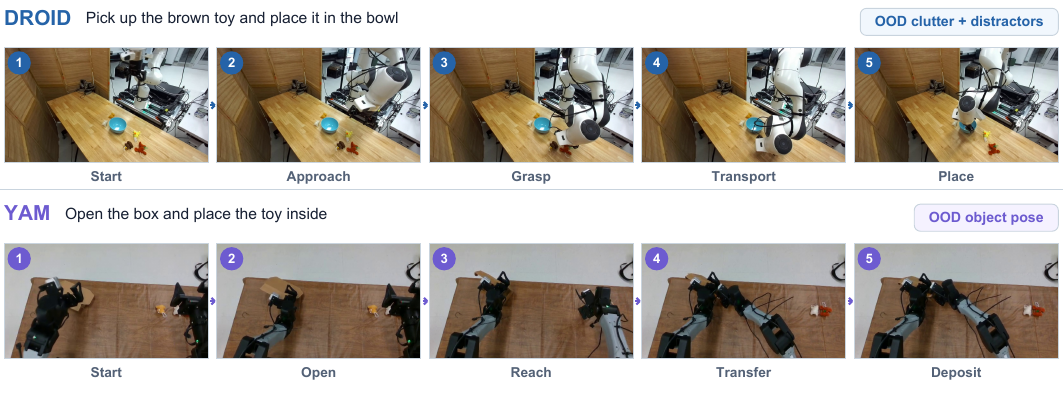}
\caption{\textbf{Successful physical-robot rollouts.} Ordered frames show DROID picking up the brown toy and placing it in the bowl (top) and YAM opening the box and placing the toy inside (bottom). Numbers indicate temporal order; circles and arrows mark the manipulated object and motion.}
\label{fig:successful-rollouts}
\end{figure*}

\begin{figure*}[!tb]
    \centering
\includegraphics[width=\linewidth,height=0.35\textheight,keepaspectratio]{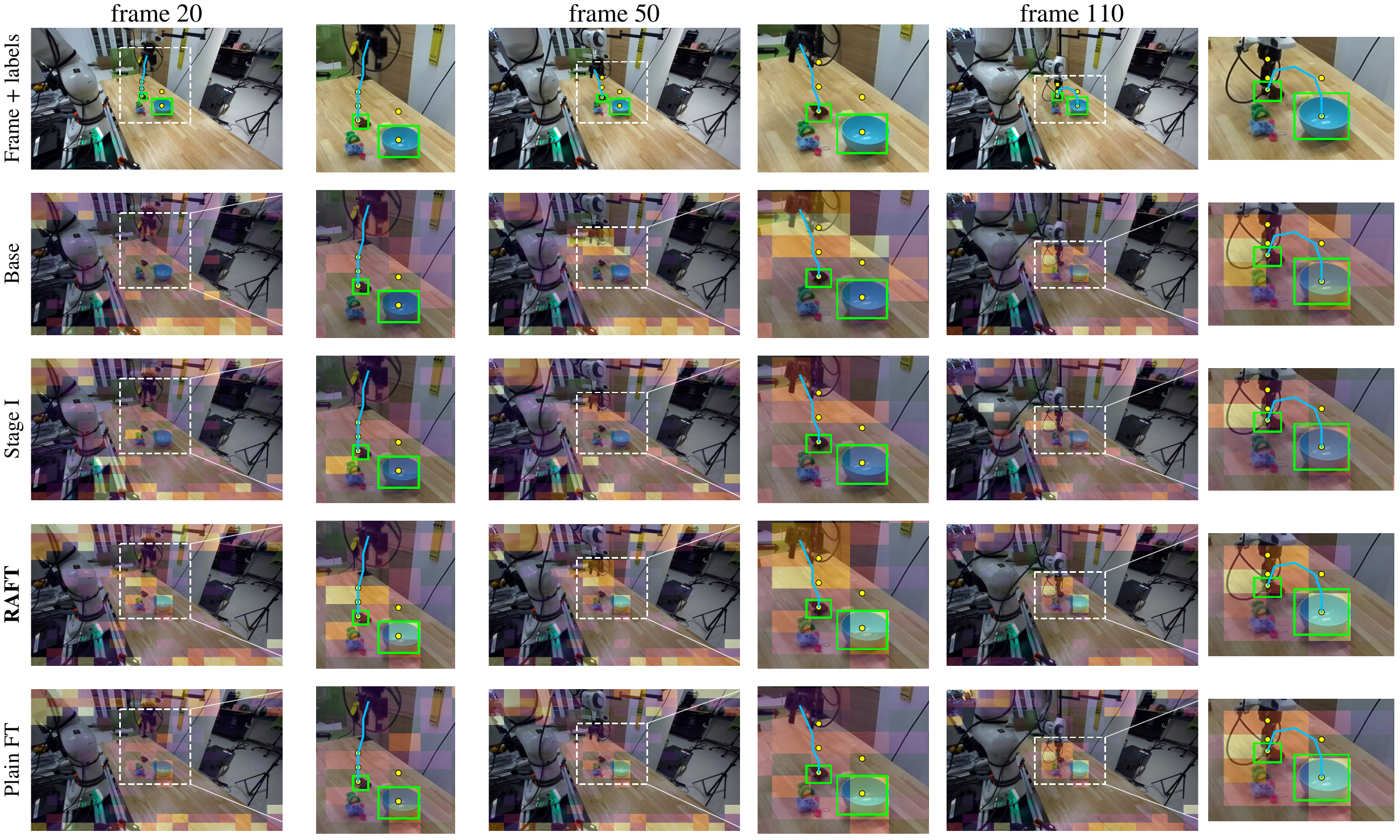}
\caption{\textbf{Attention maps along one trajectory.} Columns show successive observations; rows compare Base, Stage~I, \ours{}, and Plain FT. Overlays identify annotated task regions, and color scales are shared within each column. \ours{} places more attention on the annotated regions than Plain FT; Fig.~\ref{fig:enrichment} quantifies this across layers.}
    \label{fig:attn_map}
\end{figure*}

\begin{figure}[!ht]
  \centering
  \includegraphics[width=0.94\columnwidth]{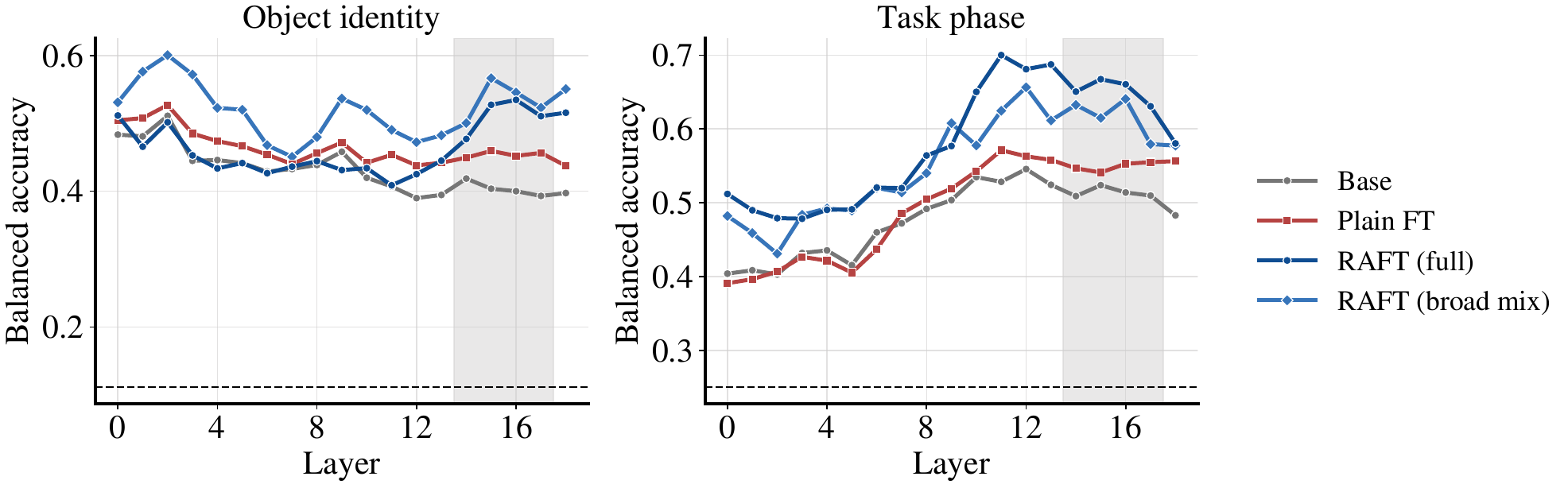}
  \caption{\textbf{Layer-wise linear probes.} Balanced accuracy for object identity (left; chance $1/7$) and episode-time quartile (right; chance $1/4$). Policies use the same 414 frames; episodes are held out from probe fitting, not policy training. Shading marks layers 14--17. In these layers, \ours{} reaches 0.51--0.53 object-identity accuracy versus 0.45 for Plain FT, and 0.62--0.65 quartile accuracy versus 0.55.}
\label{fig:late_layer_probes}
\end{figure}

\subsection{Representation Analysis}
\label{sec:diagnostics}

We examine representation dimensionality, linearly decodable task information, and visual attention to assess what \ours{} preserves during post-training. The measurement protocol is in Sec.~\ref{app:diagnostics-protocol}.

\paragraph{Late-layer dimensionality correlates with downstream success}
We measure the effective dimensionality of the backbone's hidden states with the participation ratio (PR)~\cite{Gao214262,Recanatesi2020.12.19.423618,jing2022understandingdimensionalcollapsecontrastive}:
\begin{equation}
\mathrm{PR} \;=\; \frac{\left(\sum_i \lambda_i\right)^2}{\sum_i \lambda_i^2},
\label{eq:pr}
\end{equation}
where $\lambda_i$ are the eigenvalues of the covariance of mean-pooled hidden states across frames. Across six Stage-I reasoning variants, late-layer PR correlates with success on each task-matched evaluation axis; at layer 15, absolute PR has mean Spearman $\rho=0.76$ (Fig.~\ref{fig:pr_analysis}). 

\paragraph{Object identity and task progress are more linearly decodable from \ours{}'s late layers}
We fit linear probes on the same late layers to predict object identity (seven classes, chance $1/7$) and episode-time quartile (four bins, chance $1/4$). In layers 14--17, probes on \ours{} reach 0.51--0.53 balanced accuracy for object identity, against 0.45 for Plain FT and 0.40 for the base policy, and 0.62--0.65 for episode quartile, against 0.55 and 0.51 (Fig.~\ref{fig:late_layer_probes}). Object-identity decoding also rises sharply between layers 13 and 15 for \ours{}, but barely changes for Plain FT or the base policy. The layers where PR tracks success are the same layers where this information becomes decodable.

\begin{figure*}[!tb]
  \centering
  \includegraphics[width=\linewidth]{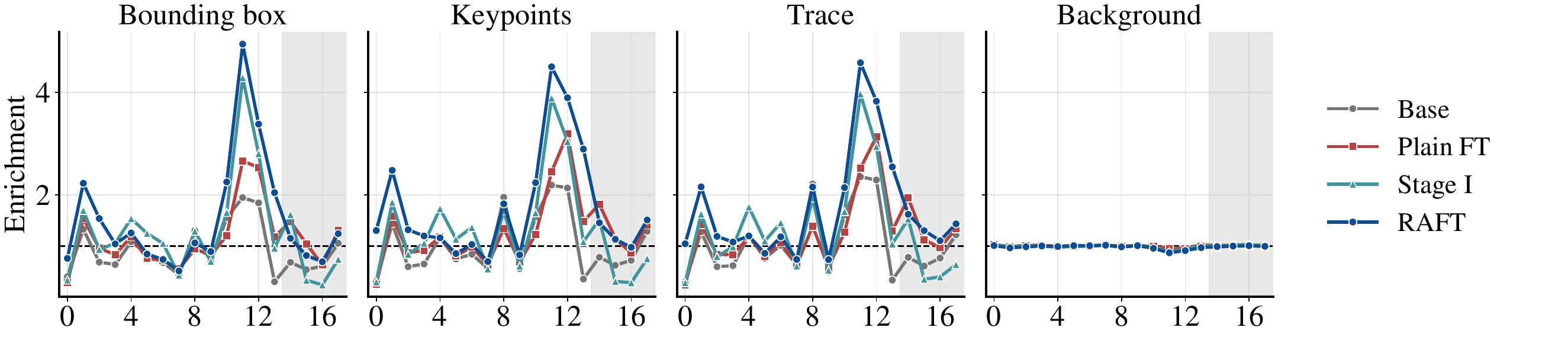}
  \caption{\textbf{Layer-wise attention enrichment.} Enrichment for object boxes, keypoints, traces, and background; values above one indicate attention exceeding a region's image-area share. Curves compare Base, Plain FT, Stage~I, and \ours{}; shading marks layers 14--17. At layer 11, \ours{} gives task regions 3.2--3.4 times their area share, versus 1.9--2.1 for Plain FT, while background attention stays near uniform.}
  \label{fig:enrichment}
\end{figure*}

\paragraph{\ours{} has higher attention enrichment on annotated task regions}
We measure how much of the policy's visual attention falls on annotated task regions, such as the manipulated object, the target, and the gripper, relative to their share of the image (enrichment; values above one mean attention is concentrated there). At layer 11, \ours{} gives these regions 3.2--3.4 times their uniform share of attention, against 1.9--2.1 for Plain FT, while attention to the background stays near uniform (Figs.~\ref{fig:attn_map} and~\ref{fig:enrichment}); the separation narrows in later layers. The concentration persists through Stage-II action training and stays above that of plain fine-tuning. The size of these effects depends on the reasoning modality used in Stage~I; Sec.~\ref{subsec:modality} reports the full modality ablation.

\begin{figure*}[!tb]
\centering
\includegraphics[width=0.27\textwidth]{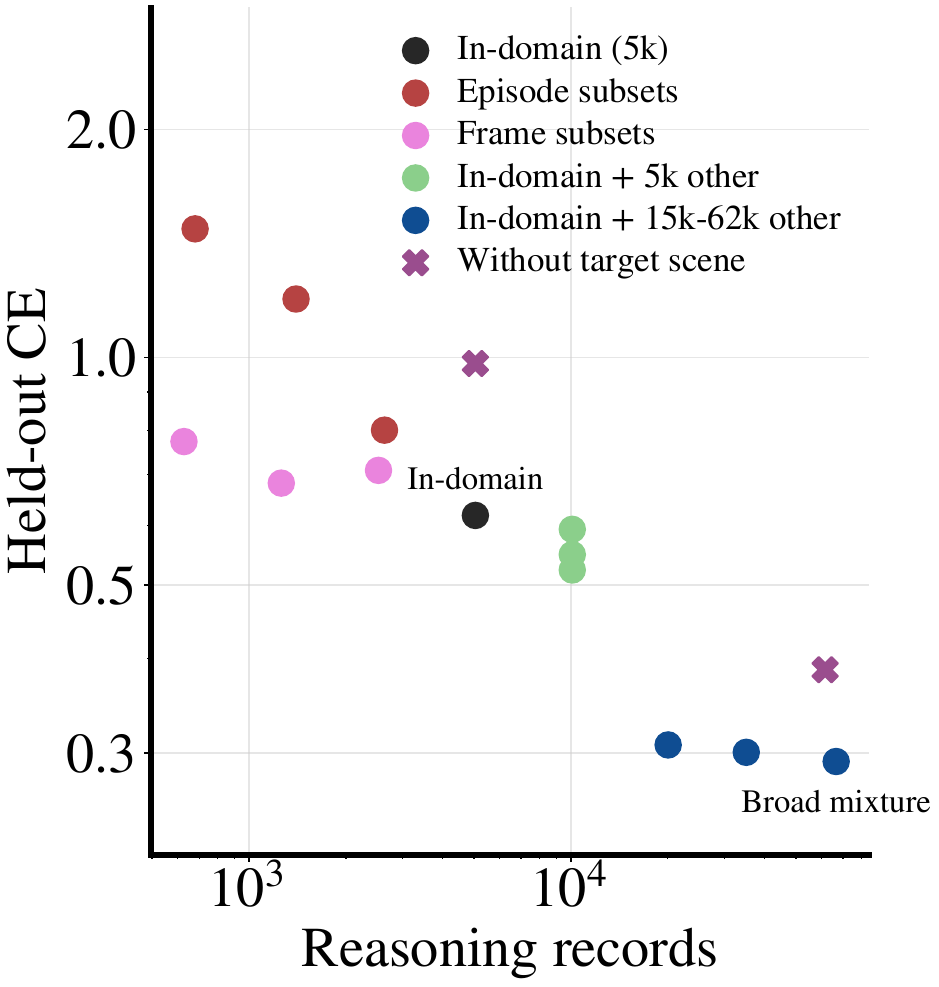}\hfill
\includegraphics[width=0.27\textwidth]{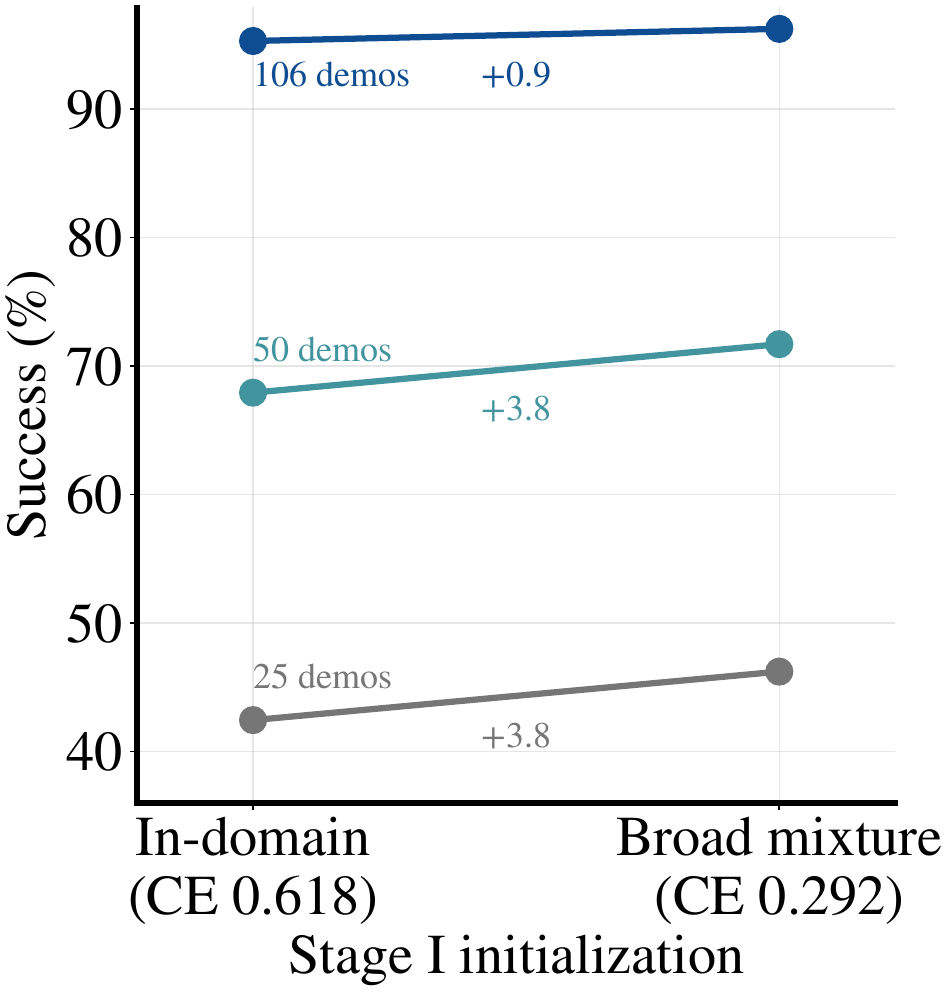}\hfill
\includegraphics[width=0.27\textwidth]{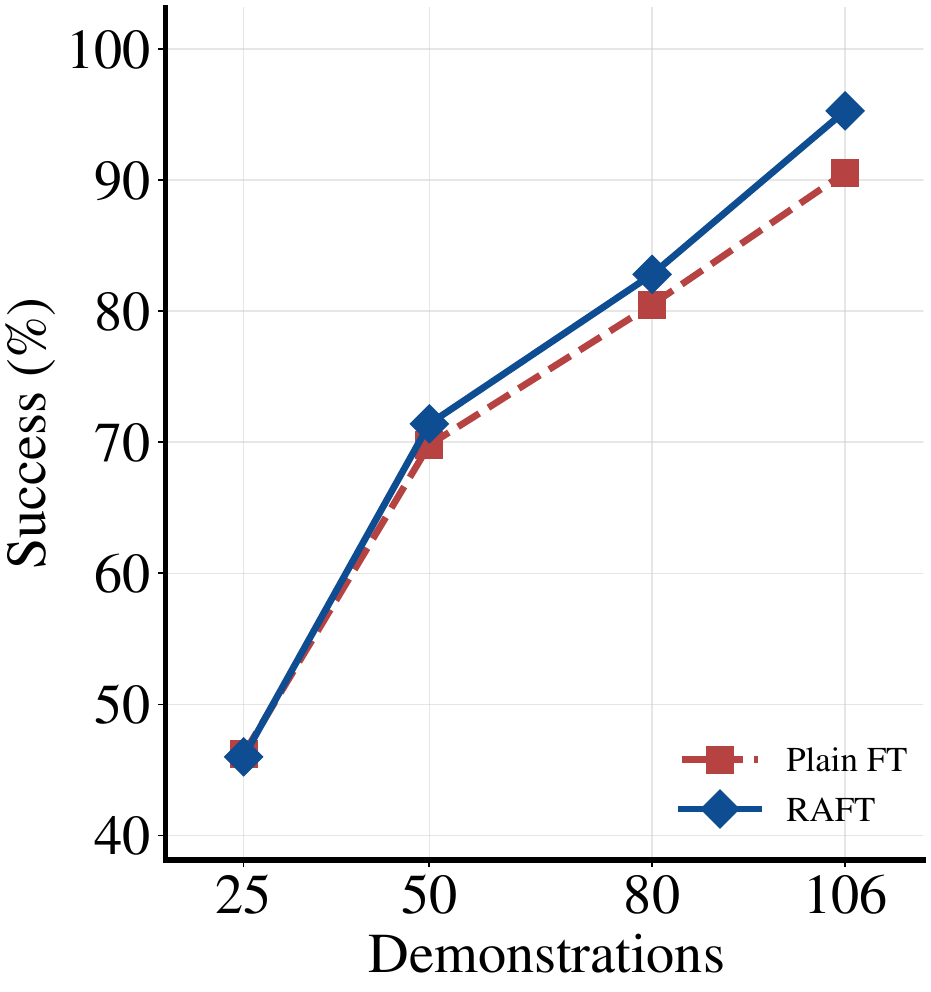}
\caption{\textbf{Effect of reasoning data and demonstration budget.} Left: held-out reasoning loss across corpus sizes and constructions. Middle: in-distribution pick-and-place success after 5{,}000 Stage-II steps from in-domain and broader-mixture Stage-I initializations. Right: \ours{} versus Plain FT across demonstration budgets, with 106 evaluation episodes per budget. The broader mixture lowers held-out loss (0.29 versus 0.62) and improves this success by 0.9--3.8 points at every budget.}
\label{fig:scaling}
\end{figure*}

\subsection{Effect of reasoning data on held-out loss and downstream success}
\label{sec:scale}

\paragraph{More reasoning data lowers held-out loss}
On a common 560-record held-out split, cross-entropy on full-reasoning targets falls as the number of reasoning records grows, in both frame- and episode-subsampled series (Fig.~\ref{fig:scaling}, left). Corpus composition also matters: 628 frame-subsampled records reach a loss of 0.77, while 680 episode-subsampled records reach only 1.48. The broader mixture, which adds roughly 60,000 records to the 5,000-record in-domain corpus, gives the lowest loss (0.29, versus 0.62 in-domain) and still reaches 0.39 with target-simulator records removed.

\paragraph{Broader reasoning data also helps downstream}
Initializing Stage~II from the broader-mixture adapter instead of the in-domain one improves in-distribution success by 0.9--3.8 points at each of the three demonstration budgets (Fig.~\ref{fig:scaling}, middle). Against Plain FT at matched budgets (right panel), the two methods tie with 25 demonstrations ($46.2\%$ success each), and \ours{} leads modestly at intermediate budgets. With all demonstrations, \ours{} reaches $95.3\%$ ID success against $90.6\%$ for Plain FT (Fig.~\ref{fig:sim_results}).

\subsection{Effect of unfrozen adapter}
\label{sec:cotrain-main}
\begin{wrapfigure}{r}{0.4\textwidth}
\vspace{-14pt}
\centering
\includegraphics[width=\linewidth]{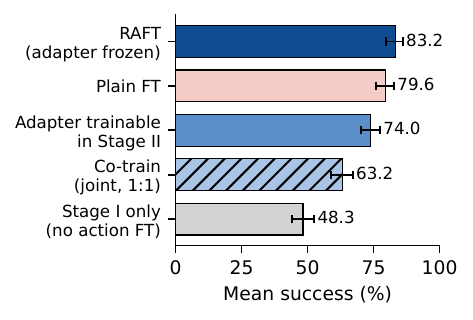}
\caption{\textbf{Training-scheme ablation on simulated pick and place.} Mean success over trials. Staged training with a frozen adapter achieves the highest success ($83.2\%$), above Plain FT ($79.6\%$), a trainable adapter ($74.0\%$), and co-training ($63.2\%$).}
\label{fig:cotrain_ablation}
\vspace{-12pt}
\end{wrapfigure}

We ablate two design choices of \ours{} on simulated pick and place: how reasoning and action supervision are combined, and whether the reasoning adapter is updated during action learning (Fig.~\ref{fig:cotrain_ablation}). \emph{Co-train} trains on reasoning and action examples jointly, sampling them equally and weighting their losses 1:1. \emph{Adapter trainable} runs both stages but keeps the Stage-I adapter trainable in Stage~II, and \emph{Stage I only} applies the adapter without action fine-tuning.

\makeatletter
\par\penalty-100\begingroup\dimen@=16\baselineskip\dimen@ii\pagegoal\advance\dimen@ii-\pagetotal
\ifdim\dimen@>\dimen@ii\ifdim\dimen@ii>\z@\vfil\fi\break\fi\endgroup
\makeatother
Across five matched splits with 106 episodes each, co-training reaches $63.21\%$ mean success, below both Plain FT ($79.62\%$) and \ours{} ($83.21\%$). Updating the adapter in Stage~II lowers success to $73.96\%$, and Stage~I alone reaches $48.30\%$. Plain FT with a frozen vision encoder also stays at $79.62\%$, so freezing the pretrained vision weights alone does not account for the gain. Split-level results are in Tables~\ref{tab:stage-ablation} and~\ref{tab:cotrain-control}.

\FloatBarrier
\section{Conclusion}

We presented \ours{}, which first trains a low-rank adapter on task-relevant reasoning annotations, freezes it, and only then fine-tunes the pretrained VLA on actions. Across two simulated tasks and two real robots, this staging improved success under distribution shift relative to plain fine-tuning, RETAIN, and DeLock. We also show how training reasoning adapter on different modalities can lead to varied performance gains and selection of optimal reasoning modality for a specific task remains an open question. Since, RAFT approaches post-training from a different perspective to current approaches, it makes RAFT compatible with these approaches and future work could explore utlizing a combination of RAFT and existing methods to do more robust post-training.

\section*{Acknowledgments}
This research was partially supported by the Toyota Research Institute and Microsoft Research, with compute support from Google TPU Research Cloud, NVIDIA Academic Grant Program, and Gemini Academic Program. The authors also thank Samuel M. Bateman and Lihan Zha for helpful discussions.

\clearpage
\bibliographystyle{plainnat}
\bibliography{citations_final}

@article{gao2026taxonomy,
  title={A taxonomy for evaluating generalist robot manipulation policies},
  author={Gao, Jensen and Belkhale, Suneel and Dasari, Sudeep and Balakrishna, Ashwin and Shah, Dhruv and Sadigh, Dorsa},
  journal={IEEE Robotics and Automation Letters},
  year={2026},
  publisher={IEEE}
}

@inproceedings{jang2024model,
  title={Model stock: All we need is just a few fine-tuned models},
  author={Jang, Dong-Hwan and Yun, Sangdoo and Han, Dongyoon},
  booktitle={European Conference on Computer Vision},
  pages={207--223},
  year={2024},
  organization={Springer}
}

@inproceedings{zhang2025effective,
  title={Effective tuning strategies for generalist robot manipulation policies},
  author={Zhang, Wenbo and Li, Yang and Qiao, Yanyuan and Huang, Siyuan and Liu, Jiajun and Dayoub, Feras and Ma, Xiao and Liu, Lingqiao},
  booktitle={2025 IEEE International Conference on Robotics and Automation (ICRA)},
  pages={7255--7262},
  year={2025},
  organization={IEEE}
}

@article{fang2026molmoact2,
  title={{MolmoAct2}: Action reasoning models for real-world deployment},
  author={Fang, Haoquan and Duan, Jiafei and Clay, Donovan and Wang, Sam and Liu, Shuo and Huang, Weikai and Fan, Xiang and Tsai, Wei-Chuan and Chen, Shirui and Wang, Yi Ru and others},
  journal={arXiv preprint arXiv:2605.02881},
  year={2026}
}

@article{wang2025generalization,
  title={Generalization capability for imitation learning},
  author={Wang, Yixiao},
  journal={arXiv preprint arXiv:2504.18538},
  year={2025}
}

@article{zha2026lap,
  title={{LAP}: Language-action pre-training enables zero-shot cross-embodiment transfer},
  author={Zha, Lihan and Hancock, Asher J and Zhang, Mingtong and Yin, Tenny and Huang, Yixuan and Shah, Dhruv and Ren, Allen Z and Majumdar, Anirudha},
  journal={arXiv preprint arXiv:2602.10556},
  year={2026}
}

@article{xiang2025parallels,
  title={Parallels between vla model post-training and human motor learning: Progress, challenges, and trends},
  author={Xiang, Tian-Yu and Jin, Ao-Qun and Zhou, Xiao-Hu and Gui, Mei-Jiang and Xie, Xiao-Liang and Liu, Shi-Qi and Wang, Shuang-Yi and Duan, Sheng-Bin and Xie, Fu-Chao and Wang, Wen-Kai and others},
  journal={arXiv preprint arXiv:2506.20966},
  year={2025}
}

@article{kaplanis2019policy,
  title={Policy consolidation for continual reinforcement learning},
  author={Kaplanis, Christos and Shanahan, Murray and Clopath, Claudia},
  journal={arXiv preprint arXiv:1902.00255},
  year={2019}
}

@article{kim2025fine,
  title={Fine-tuning vision-language-action models: Optimizing speed and success},
  author={Kim, Moo Jin and Finn, Chelsea and Liang, Percy},
  journal={arXiv preprint arXiv:2502.19645},
  year={2025}
}

@article{pertsch2025fast,
  title={{FAST}: Efficient action tokenization for vision-language-action models},
  author={Pertsch, Karl and Stachowicz, Kyle and Ichter, Brian and Driess, Danny and Nair, Suraj and Vuong, Quan and Mees, Oier and Finn, Chelsea and Levine, Sergey},
  journal={arXiv preprint arXiv:2501.09747},
  year={2025}
}

@article{kim2024openvla,
  title={{OpenVLA}: An open-source vision-language-action model},
  author={Kim, Moo Jin and Pertsch, Karl and Karamcheti, Siddharth and Xiao, Ted and Balakrishna, Ashwin and Nair, Suraj and Rafailov, Rafael and Foster, Ethan and Lam, Grace and Sanketi, Pannag and others},
  journal={arXiv preprint arXiv:2406.09246},
  year={2024}
}

@article{deshpande2026molmob0t,
  title={{MolmoB0T}: Large-Scale Simulation Enables Zero-Shot Manipulation},
  author={Deshpande, Abhay and Guru, Maya and Hendrix, Rose and Jauhri, Snehal and Eftekhar, Ainaz and Tripathi, Rohun and Argus, Max and Salvador, Jordi and Fang, Haoquan and Wallingford, Matthew and others},
  journal={arXiv preprint arXiv:2603.16861},
  year={2026}
}

@article{bjorck2025gr00t,
  title={{GR00T} {N1}: An open foundation model for generalist humanoid robots},
  author={Bjorck, Johan and Casta{\~n}eda, Fernando and Cherniadev, Nikita and Da, Xingye and Ding, Runyu and Fan, Linxi and Fang, Yu and Fox, Dieter and Hu, Fengyuan and Huang, Spencer and others},
  journal={arXiv preprint arXiv:2503.14734},
  year={2025}
}

@article{black2024pi_0,
  title={{$\pi_0$}: A Vision-Language-Action Flow Model for General Robot Control},
  author={Black, Kevin and Brown, Noah and Driess, Danny and Esmail, Adnan and Equi, Michael and Finn, Chelsea and Fusai, Niccolo and Groom, Lachy and Hausman, Karol and Ichter, Brian and others},
  journal={arXiv preprint arXiv:2410.24164},
  year={2024}
}

@article{khazatsky2024droid,
  title={{DROID}: A large-scale in-the-wild robot manipulation dataset},
  author={Khazatsky, Alexander and Pertsch, Karl and Nair, Suraj and Balakrishna, Ashwin and Dasari, Sudeep and Karamcheti, Siddharth and Nasiriany, Soroush and Srirama, Mohan Kumar and Chen, Lawrence Yunliang and Ellis, Kirsty and others},
  journal={arXiv preprint arXiv:2403.12945},
  year={2024}
}

@article{bousmalis2023robocat,
  title={Robocat: A self-improving generalist agent for robotic manipulation},
  author={Bousmalis, Konstantinos and Vezzani, Giulia and Rao, Dushyant and Devin, Coline and Lee, Alex X and Bauz{\'a}, Maria and Davchev, Todor and Zhou, Yuxiang and Gupta, Agrim and Raju, Akhil and others},
  journal={arXiv preprint arXiv:2306.11706},
  year={2023}
}

@article{li2025controlvla,
  title={{ControlVLA}: Few-shot object-centric adaptation for pre-trained vision-language-action models},
  author={Li, Puhao and Wu, Yingying and Xi, Ziheng and Li, Wanlin and Huang, Yuzhe and Zhang, Zhiyuan and Chen, Yinghan and Wang, Jianan and Zhu, Song-Chun and Liu, Tengyu and others},
  journal={arXiv preprint arXiv:2506.16211},
  year={2025}
}

@inproceedings{guo2025improving,
  title={Improving vision-language-action model with online reinforcement learning},
  author={Guo, Yanjiang and Zhang, Jianke and Chen, Xiaoyu and Ji, Xiang and Wang, Yen-Jen and Hu, Yucheng and Chen, Jianyu},
  booktitle={2025 IEEE International Conference on Robotics and Automation (ICRA)},
  pages={15665--15672},
  year={2025},
  organization={IEEE}
}

@inproceedings{wortsman2022robust,
  title={Robust fine-tuning of zero-shot models},
  author={Wortsman, Mitchell and Ilharco, Gabriel and Kim, Jong Wook and Li, Mike and Kornblith, Simon and Roelofs, Rebecca and Lopes, Raphael Gontijo and Hajishirzi, Hannaneh and Farhadi, Ali and Namkoong, Hongseok and others},
  booktitle={2022 IEEE/CVF Conference on Computer Vision and Pattern Recognition (CVPR)},
  pages={7949--7961},
  year={2022},
  organization={IEEE}
}

@article{zhou2025libero,
  title={{LIBERO-Pro}: Towards robust and fair evaluation of vision-language-action models beyond memorization},
  author={Zhou, Xueyang and Xu, Yangming and Tie, Guiyao and Chen, Yongchao and Zhang, Guowen and Chu, Duanfeng and Zhou, Pan and Sun, Lichao},
  journal={arXiv preprint arXiv:2510.03827},
  year={2025}
}

@inproceedings{fei2026libero,
  title={{LIBERO-Plus}: A progressive robustness benchmark for visual-language-action models},
  author={Fei, Senyu and Wang, Siyin and Shi, Junhao and Dai, Zihao and Cai, Jikun and Qian, Pengfang and Ji, Li and He, Xinzhe and Zhang, Shiduo and Fei, Zhaoye and others},
  booktitle={Proceedings of the IEEE/CVF Conference on Computer Vision and Pattern Recognition},
  pages={38574--38583},
  year={2026}
}

@article{zang2025rlinf,
  title={{RLinf-VLA}: A Unified and Efficient Framework for Reinforcement Learning of Vision-Language-Action Models},
  author={Zang, Hongzhi and Wei, Mingjie and Xu, Si and Wu, Yongji and Guo, Zhen and Wang, Yuanqing and Lin, Hao and Wang, Peihong and Shi, Liangzhi and Xie, Yuqing and others},
  journal={arXiv preprint arXiv:2510.06710},
  year={2025}
}

@article{liu2026vls,
  title={{VLS}: Steering pretrained robot policies via vision-language models},
  author={Liu, Shuo and Singh, Ishneet Sukhvinder and Xu, Yiqing and Duan, Jiafei and Krishna, Ranjay},
  journal={arXiv preprint arXiv:2602.03973},
  year={2026}
}

@article{chen2026steerable,
  title={Steerable vision-language-action policies for embodied reasoning and hierarchical control},
  author={Chen, William and Bhatia, Jagdeep Singh and Glossop, Catherine and Mathihalli, Nikhil and Doshi, Ria and Tang, Andy and Driess, Danny and Pertsch, Karl and Levine, Sergey},
  journal={arXiv preprint arXiv:2602.13193},
  year={2026}
}

@article{grover2025enhancing,
  title={Enhancing generalization in vision-language-action models by preserving pretrained representations},
  author={Grover, Shresth and Gopalkrishnan, Akshay and Ai, Bo and Christensen, Henrik I and Su, Hao and Li, Xuanlin},
  journal={arXiv preprint arXiv:2509.11417},
  year={2025}
}

@article{team2024octo,
  title={{Octo}: An open-source generalist robot policy},
  author={Team, Octo Model and Ghosh, Dibya and Walke, Homer and Pertsch, Karl and Black, Kevin and Mees, Oier and Dasari, Sudeep and Hejna, Joey and Kreiman, Tobias and Xu, Charles and others},
  journal={arXiv preprint arXiv:2405.12213},
  year={2024}
}

@article{sharma2023lossless,
  title={Lossless adaptation of pretrained vision models for robotic manipulation},
  author={Sharma, Mohit and Fantacci, Claudio and Zhou, Yuxiang and Koppula, Skanda and Heess, Nicolas and Scholz, Jon and Aytar, Yusuf},
  journal={arXiv preprint arXiv:2304.06600},
  year={2023}
}

@article{pan2026sop,
  title={{SOP}: A scalable online post-training system for vision-language-action models},
  author={Pan, Mingjie and Feng, Siyuan and Zhang, Qinglin and Li, Xinchen and Song, Jianheng and Qu, Chendi and Wang, Yi and Li, Chuankang and Xiong, Ziyu and Chen, Zhi and others},
  journal={arXiv preprint arXiv:2601.03044},
  year={2026}
}

@article{xing2025shortcut,
  title={Shortcut learning in generalist robot policies: The role of dataset diversity and fragmentation},
  author={Xing, Youguang and Luo, Xu and Xie, Junlin and Gao, Lianli and Shen, Hengtao and Song, Jingkuan},
  journal={arXiv preprint arXiv:2508.06426},
  year={2025}
}

@article{kim2026molmospaces,
  title={{MolmoSpaces}: A large-scale open ecosystem for robot navigation and manipulation},
  author={Kim, Yejin and Pumacay, Wilbert and Rayyan, Omar and Argus, Max and Han, Winson and VanderBilt, Eli and Salvador, Jordi and Deshpande, Abhay and Hendrix, Rose and Jauhri, Snehal and others},
  journal={arXiv preprint arXiv:2602.11337},
  year={2026}
}

@InProceedings{pmlr-v305-black25a,
  title = 	 {$\pi_{0.5}$: a Vision-Language-Action Model with Open-World Generalization},
  author =       {Black, Kevin and Brown, Noah and Darpinian, James and Dhabalia, Karan and Driess, Danny and Esmail, Adnan and Equi, Michael Robert and Finn, Chelsea and Fusai, Niccolo and Galliker, Manuel Y. and Ghosh, Dibya and Groom, Lachy and Hausman, Karol and Ichter, Brian and Jakubczak, Szymon and Jones, Tim and Ke, Liyiming and LeBlanc, Devin and Levine, Sergey and Li-Bell, Adrian and Mothukuri, Mohith and Nair, Suraj and Pertsch, Karl and Ren, Allen Z. and Shi, Lucy Xiaoyang and Smith, Laura and Springenberg, Jost Tobias and Stachowicz, Kyle and Tanner, James and Vuong, Quan and Walke, Homer and Walling, Anna and Wang, Haohuan and Yu, Lili and Zhilinsky, Ury},
  booktitle = 	 {Proceedings of The 9th Conference on Robot Learning},
  pages = 	 {17--40},
  year = 	 {2025},
  editor = 	 {Lim, Joseph and Song, Shuran and Park, Hae-Won},
  volume = 	 {305},
  series = 	 {Proceedings of Machine Learning Research},
  month = 	 {27--30 Sep},
  publisher =    {PMLR},
  url = 	 {https://proceedings.mlr.press/v305/black25a.html}
}

@article{anil2023palm,
  title={{PaLM} 2 Technical Report},
  author={Anil, Rohan and Dai, Andrew M and Firat, Orhan and Johnson, Melvin and Lepikhin, Dmitry and Passos, Alexandre and Shakeri, Siamak and Taropa, Emanuel and Bailey, Paige and Chen, Zhifeng and others},
  journal={arXiv preprint arXiv:2305.10403},
  year={2023}
}

@inproceedings{yadav2025retain,
  title={Robust Fine-Tuning of Vision-Language-Action Robot Policies via Parameter Merging},
  author={Yadav, Yajat and Zhou, Zhiyuan and Wagenmaker, Andrew and Pertsch, Karl and Levine, Sergey},
  booktitle={International Conference on Learning Representations (ICLR)},
  year={2026}
}

@article{kachaev2025dontblind,
  title={Don't Blind Your {VLA}: Aligning Visual Representations for {OOD} Generalization},
  author={Kachaev, Nikita and Kolosov, Mikhail and Zelezetsky, Daniil and Kovalev, Alexey K and Panov, Aleksandr I},
  journal={arXiv preprint arXiv:2510.25616},
  year={2025}
}

@article{huang2026breakinglockin,
  title={Breaking Lock-In: Preserving Steerability under Low-Data {VLA} Post-Training},
  author={Suning Huang and Jiaqi Shao and Ke Wang and Qianzhong Chen and Jiankai Sun and Yanjiang Guo and Mac Schwager and Jeannette Bohg},
  journal={arXiv preprint arXiv:2604.23121},
  year={2026}
}

@inproceedings{zawalski2024ecot,
  title={Robotic Control via Embodied Chain-of-Thought Reasoning},
  author={Zawalski, Micha{\l} and Chen, William and Pertsch, Karl and Mees, Oier and Finn, Chelsea and Levine, Sergey},
  booktitle={Conference on Robot Learning (CoRL)},
  year={2024}
}

@inproceedings{li2025spatial,
  title={Spatial Forcing: Implicit Spatial Representation Alignment for Vision-Language-Action Model},
  author={Li, Fuhao and Song, Wenxuan and Zhao, Han and Wang, Jingbo and Ding, Pengxiang and Wang, Donglin and Zeng, Long and Li, Haoang},
  booktitle={International Conference on Learning Representations (ICLR)},
  year={2026}
}

@inproceedings{li2025simplevla,
  title={{SimpleVLA-RL}: Scaling {VLA} Training via Reinforcement Learning},
  author={Li, Haozhan and Zuo, Yuxin and Yu, Jiale and Zhang, Yuhao and Yang, Zhaohui and Zhang, Kaiyan and Zhu, Xuekai and Zhang, Yuchen and Chen, Tianxing and Cui, Ganqu and others},
  booktitle={International Conference on Learning Representations (ICLR)},
  year={2026}
}

@inproceedings{dey2024revla,
  title={{ReVLA}: Reverting Visual Domain Limitation of Robotic Foundation Models},
  author={Dey, Sombit and Zaech, Jan-Nico and Nikolov, Nikolay and Van Gool, Luc and Paudel, Danda Pani},
  booktitle={IEEE International Conference on Robotics and Automation (ICRA)},
  year={2025}
}

@inproceedings{
hu2022lora,
title={Lo{RA}: Low-Rank Adaptation of Large Language Models},
author={Edward J Hu and Yelong Shen and Phillip Wallis and Zeyuan Allen-Zhu and Yuanzhi Li and Shean Wang and Lu Wang and Weizhu Chen},
booktitle={International Conference on Learning Representations},
year={2022},
url={https://openreview.net/forum?id=nZeVKeeFYf9}
}

@article{Gao214262,
  title={A theory of multineuronal dimensionality, dynamics and measurement},
  author={Gao, Peiran and Trautmann, Eric and Yu, Byron and Santhanam, Gopal and Ryu, Stephen and Shenoy, Krishna and Ganguli, Surya},
  journal={bioRxiv},
  year={2017},
  doi={10.1101/214262}
}

@article{Recanatesi2020.12.19.423618,
  title={A scale-dependent measure of system dimensionality},
  author={Recanatesi, Stefano and Bradde, Serena and Balasubramanian, Vijay and Steinmetz, Nicholas A. and Shea-Brown, Eric},
  journal={Patterns},
  volume={3},
  number={8},
  pages={100555},
  year={2022},
  doi={10.1016/j.patter.2022.100555}
}

@inproceedings{jing2022understandingdimensionalcollapsecontrastive,
  title={Understanding Dimensional Collapse in Contrastive Self-supervised Learning},
  author={Li Jing and Pascal Vincent and Yann LeCun and Yuandong Tian},
  booktitle={International Conference on Learning Representations (ICLR)},
  year={2022}
}

@misc{open_x_embodiment_rt_x_2023,
title={Open {X-E}mbodiment: Robotic Learning Datasets and {RT-X} Models},
author = {Open X-Embodiment Collaboration and Abby O'Neill and Abdul Rehman and Abhinav Gupta and Abhiram Maddukuri and Abhishek Gupta and Abhishek Padalkar and Abraham Lee and Acorn Pooley and Agrim Gupta and Ajay Mandlekar and Ajinkya Jain and Albert Tung and Alex Bewley and Alex Herzog and Alex Irpan and Alexander Khazatsky and Anant Rai and Anchit Gupta and Andrew Wang and Andrey Kolobov and Anikait Singh and Animesh Garg and Aniruddha Kembhavi and Annie Xie and Anthony Brohan and Antonin Raffin and Archit Sharma and Arefeh Yavary and Arhan Jain and Ashwin Balakrishna and Ayzaan Wahid and Ben Burgess-Limerick and Beomjoon Kim and Bernhard Schölkopf and Blake Wulfe and Brian Ichter and Cewu Lu and Charles Xu and Charlotte Le and Chelsea Finn and Chen Wang and Chenfeng Xu and Cheng Chi and Chenguang Huang and Christine Chan and Christopher Agia and Chuer Pan and Chuyuan Fu and Coline Devin and Danfei Xu and Daniel Morton and Danny Driess and Daphne Chen and Deepak Pathak and Dhruv Shah and Dieter Büchler and Dinesh Jayaraman and Dmitry Kalashnikov and Dorsa Sadigh and Edward Johns and Ethan Foster and Fangchen Liu and Federico Ceola and Fei Xia and Feiyu Zhao and Felipe Vieira Frujeri and Freek Stulp and Gaoyue Zhou and Gaurav S. Sukhatme and Gautam Salhotra and Ge Yan and Gilbert Feng and Giulio Schiavi and Glen Berseth and Gregory Kahn and Guangwen Yang and Guanzhi Wang and Hao Su and Hao-Shu Fang and Haochen Shi and Henghui Bao and Heni Ben Amor and Henrik I Christensen and Hiroki Furuta and Homanga Bharadhwaj and Homer Walke and Hongjie Fang and Huy Ha and Igor Mordatch and Ilija Radosavovic and Isabel Leal and Jacky Liang and Jad Abou-Chakra and Jaehyung Kim and Jaimyn Drake and Jan Peters and Jan Schneider and Jasmine Hsu and Jay Vakil and Jeannette Bohg and Jeffrey Bingham and Jeffrey Wu and Jensen Gao and Jiaheng Hu and Jiajun Wu and Jialin Wu and Jiankai Sun and Jianlan Luo and Jiayuan Gu and Jie Tan and Jihoon Oh and Jimmy Wu and Jingpei Lu and Jingyun Yang and Jitendra Malik and João Silvério and Joey Hejna and Jonathan Booher and Jonathan Tompson and Jonathan Yang and Jordi Salvador and Joseph J. Lim and Junhyek Han and Kaiyuan Wang and Kanishka Rao and Karl Pertsch and Karol Hausman and Keegan Go and Keerthana Gopalakrishnan and Ken Goldberg and Kendra Byrne and Kenneth Oslund and Kento Kawaharazuka and Kevin Black and Kevin Lin and Kevin Zhang and Kiana Ehsani and Kiran Lekkala and Kirsty Ellis and Krishan Rana and Krishnan Srinivasan and Kuan Fang and Kunal Pratap Singh and Kuo-Hao Zeng and Kyle Hatch and Kyle Hsu and Laurent Itti and Lawrence Yunliang Chen and Lerrel Pinto and Li Fei-Fei and Liam Tan and Linxi "Jim" Fan and Lionel Ott and Lisa Lee and Luca Weihs and Magnum Chen and Marion Lepert and Marius Memmel and Masayoshi Tomizuka and Masha Itkina and Mateo Guaman Castro and Max Spero and Maximilian Du and Michael Ahn and Michael C. Yip and Mingtong Zhang and Mingyu Ding and Minho Heo and Mohan Kumar Srirama and Mohit Sharma and Moo Jin Kim and Muhammad Zubair Irshad and Naoaki Kanazawa and Nicklas Hansen and Nicolas Heess and Nikhil J Joshi and Niko Suenderhauf and Ning Liu and Norman Di Palo and Nur Muhammad Mahi Shafiullah and Oier Mees and Oliver Kroemer and Osbert Bastani and Pannag R Sanketi and Patrick "Tree" Miller and Patrick Yin and Paul Wohlhart and Peng Xu and Peter David Fagan and Peter Mitrano and Pierre Sermanet and Pieter Abbeel and Priya Sundaresan and Qiuyu Chen and Quan Vuong and Rafael Rafailov and Ran Tian and Ria Doshi and Roberto Mart{\'i}n-Mart{\'i}n and Rohan Baijal and Rosario Scalise and Rose Hendrix and Roy Lin and Runjia Qian and Ruohan Zhang and Russell Mendonca and Rutav Shah and Ryan Hoque and Ryan Julian and Samuel Bustamante and Sean Kirmani and Sergey Levine and Shan Lin and Sherry Moore and Shikhar Bahl and Shivin Dass and Shubham Sonawani and Shubham Tulsiani and Shuran Song and Sichun Xu and Siddhant Haldar and Siddharth Karamcheti and Simeon Adebola and Simon Guist and Soroush Nasiriany and Stefan Schaal and Stefan Welker and Stephen Tian and Subramanian Ramamoorthy and Sudeep Dasari and Suneel Belkhale and Sungjae Park and Suraj Nair and Suvir Mirchandani and Takayuki Osa and Tanmay Gupta and Tatsuya Harada and Tatsuya Matsushima and Ted Xiao and Thomas Kollar and Tianhe Yu and Tianli Ding and Todor Davchev and Tony Z. Zhao and Travis Armstrong and Trevor Darrell and Trinity Chung and Vidhi Jain and Vikash Kumar and Vincent Vanhoucke and Vitor Guizilini and Wei Zhan and Wenxuan Zhou and Wolfram Burgard and Xi Chen and Xiangyu Chen and Xiaolong Wang and Xinghao Zhu and Xinyang Geng and Xiyuan Liu and Xu Liangwei and Xuanlin Li and Yansong Pang and Yao Lu and Yecheng Jason Ma and Yejin Kim and Yevgen Chebotar and Yifan Zhou and Yifeng Zhu and Yilin Wu and Ying Xu and Yixuan Wang and Yonatan Bisk and Yongqiang Dou and Yoonyoung Cho and Youngwoon Lee and Yuchen Cui and Yue Cao and Yueh-Hua Wu and Yujin Tang and Yuke Zhu and Yunchu Zhang and Yunfan Jiang and Yunshuang Li and Yunzhu Li and Yusuke Iwasawa and Yutaka Matsuo and Zehan Ma and Zhuo Xu and Zichen Jeff Cui and Zichen Zhang and Zipeng Fu and Zipeng Lin},
howpublished  = {\url{https://arxiv.org/abs/2310.08864}},
year = {2023},
}

\clearpage
\beginappendix
\section{Training and implementation details}
\label{app:training-details}

Table~\ref{tab:training-details} reports the training configuration \ours{} uses in simulation and on DROID (SigLIP vision encoder and an 18-layer Gemma-2B decoder). The YAM experiments instead adapt $\pi_{0.5}$ using a similar setup. In \ours{}, the pretrained SigLIP weights remain frozen in both stages. Stage~I optimizes reasoning LoRA modules in the language model and vision encoder. 
\begin{table*}[t]
\centering
\small
\setlength{\tabcolsep}{5pt}
\renewcommand{\arraystretch}{1.16}
\caption{\textbf{Training configuration for the two stages of \ours{}}.}
\label{tab:training-details}
\begin{tabular}{p{0.15\textwidth} p{0.39\textwidth} p{0.39\textwidth}}
\toprule
 & \textbf{Stage I: reasoning alignment} & \textbf{Stage II: action fine-tuning} \\
\midrule
Trainable components
& LoRA adapters only
& Language model, action expert, and projection layers; SigLIP vision tower and Stage-I LoRA frozen \\

LoRA placement
& All 18 language-model layers: q/k/v/o attention projections and gate/up/down MLP projections (126 modules); all 27 vision layers: q/k/v projections (81 modules)
& The Stage-I LoRA remains attached and frozen; no adapter parameters are optimized in Stage II \\

LoRA configuration
& Rank $r=64$, scaling $\alpha=128$ ($\alpha/r=2.0$), dropout $0.05$
& --- \\

Trainable parameters
& $90.4$M ($78.4$M language-model and $11.9$M vision-adapter parameters)
& $3.09$B of $3.50$B parameters \\

Optimizer
& AdamW, $\beta=(0.9,0.999)$, weight decay $10^{-6}$, gradient clipping $1.0$
& AdamW, $\beta=(0.9,0.95)$, gradient clipping $1.0$ \\

Learning rate
& Peak $2\times10^{-4}$; 100 warmup steps; cosine schedule (2,010-step horizon), 1,000 steps
& Peak $5\times10^{-5}$; 1,000 warmup steps; thereafter effectively constant at $5\times10^{-5}$ \\

Effective batch size
& $12$ samples/GPU $\times$ accumulation $4 = 48$
& $16$ samples/GPU $\times$ 4 GPUs $\times$ accumulation $8 = 512$ \\

Optimization steps
& 1,000
& 5,000 \\

Numerics
& bfloat16 with gradient checkpointing
& bfloat16 with gradient checkpointing \\

Hardware
& 1$\times$ NVIDIA L40 (46 GB)
& 4$\times$ NVIDIA L40 (46 GB), one node with distributed data parallel training \\
\bottomrule
\end{tabular}
\end{table*}

\subsection{Annotation format}
\label{app:gemini-format}

For each observation and task instruction, Gemini~3.1~Pro is asked to return a JSON object in the following format. Coordinates are normalized to the $[0,1000]$ image frame; boxes are $[x_1, y_1, x_2, y_2]$ and points are $(x, y)$. The format below is for pick-and-place; the open drawer task uses the same structure with \texttt{handle\_point}, \texttt{pre\_grasp\_point}, \texttt{grasp\_point}, \texttt{pull\_start\_point}, and \texttt{pull\_end\_point} as keypoints.

\begin{verbatim}
{
  "reasoning": "<one or two sentences about relevant objects and positions>",
  "objects": [
    {"name": "<label>", "bbox": [x1, y1, x2, y2]}
  ],
  "keypoints": {
    "grasp_point":     [x, y],
    "target_point":    [x, y],
    "lift_point":      [x, y],
    "pre_grasp_point": [x, y],
    "pre_place_point": [x, y]
  },
  "trace": [[x, y], ...]
}
\end{verbatim}

\paragraph{Bimanual (YAM).} For the bimanual box-loading task, one arm holds the lid open while the other loads the object, so the keypoints are split by arm rather than by motion stage, and the response adds a per-arm block:

\begin{verbatim}
{
  "reasoning": "<one or two sentences about relevant objects and positions>",
  "objects": [
    {"name": "<label>", "bbox": [x1, y1, x2, y2]}
  ],
  "keypoints": {
    "left_ee_point":  [x, y],
    "lid_point":      [x, y],
    "right_ee_point": [x, y],
    "object_point":   [x, y],
    "drop_point":     [x, y]
  },
  "trace": [[x, y], ...],
  "arms": {
    "left":  {"motion": "<what the LEFT arm is doing in this frame>",
              "trace": [[x, y], ...]},
    "right": {"motion": "<what the RIGHT arm is doing in this frame>",
              "trace": [[x, y], ...]}
  }
}
\end{verbatim}

\subsection{Stage-I supervision targets by modality}
\label{app:stage1-modality-examples}

Table~\ref{tab:stage1-modality-examples} illustrates the target formats for a mug-into-bowl pick-and-place example. Every row uses the same pick-and-place observation and instruction; only the response format changes. VQA and spatial supervision contain language only, while geometric variants use the corresponding coordinates and, where applicable, object or keypoint labels.

\begin{table*}[t]
\centering
\small
\setlength{\tabcolsep}{6pt}
\renewcommand{\arraystretch}{1.22}
\caption{\textbf{Modality-specific Stage-I targets for a mug-into-bowl pick-and-place example.} Coordinates use the normalized $[0,1000]$ image frame.}
\label{tab:stage1-modality-examples}
\begin{tabular}{p{0.15\textwidth} p{0.80\textwidth}}
\toprule
\textbf{Variant} & \textbf{Target response} \\
\midrule
VQA
& The robot should grasp the glossy beige ceramic mug on the left side of the counter and place it into the white bowl on its right. \\

Spatial
& The beige mug is left of the white bowl. The gripper begins above and to the right of the mug, moves to the mug, lifts it, then moves rightward to the bowl. \\

Bounding box
& \texttt{glossy beige ceramic mug: [310, 600, 380, 740]; white bowl: [410, 570, 520, 720]} \\

Keypoints
& \texttt{grasp\_point: [340, 670]; target\_point: [460, 650]; lift\_point: [340, 500]; pre\_grasp\_point: [340, 550]; pre\_place\_point: [460, 500]} \\

Trajectory trace
& \texttt{[[600, 400], [470, 470], [340, 550], [340, 670], [400, 580], [460, 650]]} \\

Full reasoning
& The VQA plan together with the two object boxes, five semantic keypoints, and six-point trajectory trace shown above. \\
\bottomrule
\end{tabular}
\end{table*}

\subsection{Baseline implementation details}
\label{app:baseline-details}

\textbf{Plain FT.} Plain fine-tuning optimizes the SigLIP vision encoder, language model, action expert, and projection layers without Stage-I reasoning alignment.

\textbf{RETAIN and DeLock.} We follow the methods as defined in their original papers~\cite{yadav2025retain,huang2026breakinglockin}. For our evaluations, RETAIN uses the merge coefficient $\alpha=0.9$ reported by its authors. Both baselines use trainable SigLIP weights during fine-tuning. DeLock regularizes the drift of those weights from their pretrained values with coefficient $10^{-2}$. We tune its contrastive-prompt coefficient over the values described below.

\textbf{Contrastive prompt.} We sweep the contrastive-prompt coefficient over $\{0.5, 2, 3\}$ and report the setting with the highest average success across evaluation axes.

\subsection{Evaluation protocol}
\label{app:evaluation-protocol}

Simulation task definitions, reset procedure, matched initial states, and success criteria follow the MolmoSpaces evaluation protocol~\cite{kim2026molmospaces}. Within each evaluation axis, methods are compared on matched episode IDs. Physical-robot resets preserve the task configuration and start state. DROID used 30 ID trials and 24, 24, and 20 trials for the object, distractor, and background shifts; YAM used 30 trials per OOD axis.
\section{Qualitative real-robot comparisons}
\label{app:qualitative-rollouts}

\begin{figure*}[p]
\centering
\includegraphics[width=\textwidth]{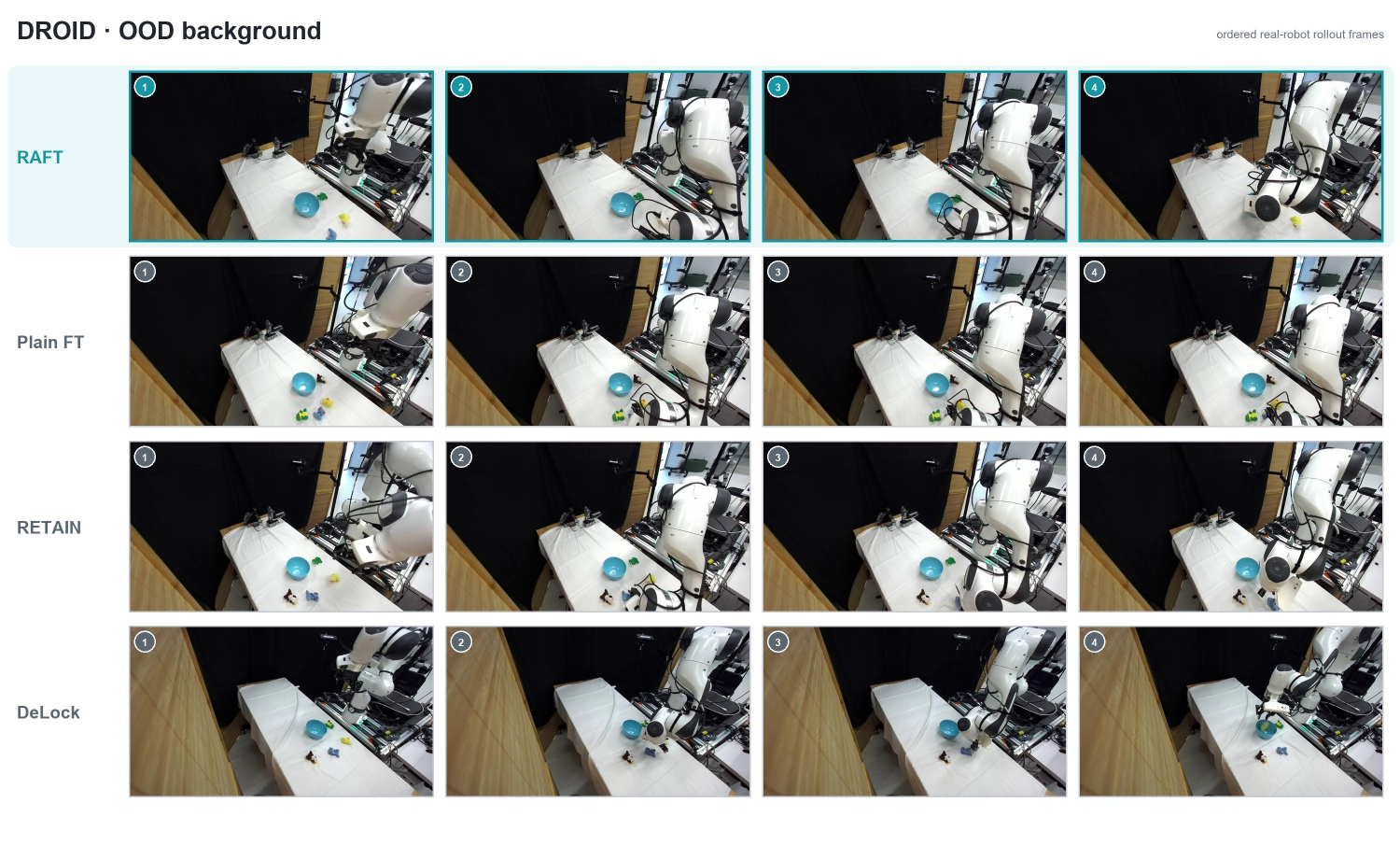}
\caption{\textbf{DROID rollout comparison under an OOD background.} Rows show \ours{}, Plain FT, RETAIN, and DeLock at four ordered points in matched real-robot episodes. \ours{} completes the manipulation, while the comparison policies fail before task completion.}
\label{fig:droid-rollout-background}
\end{figure*}

\begin{figure*}[p]
\centering
\includegraphics[width=\textwidth]{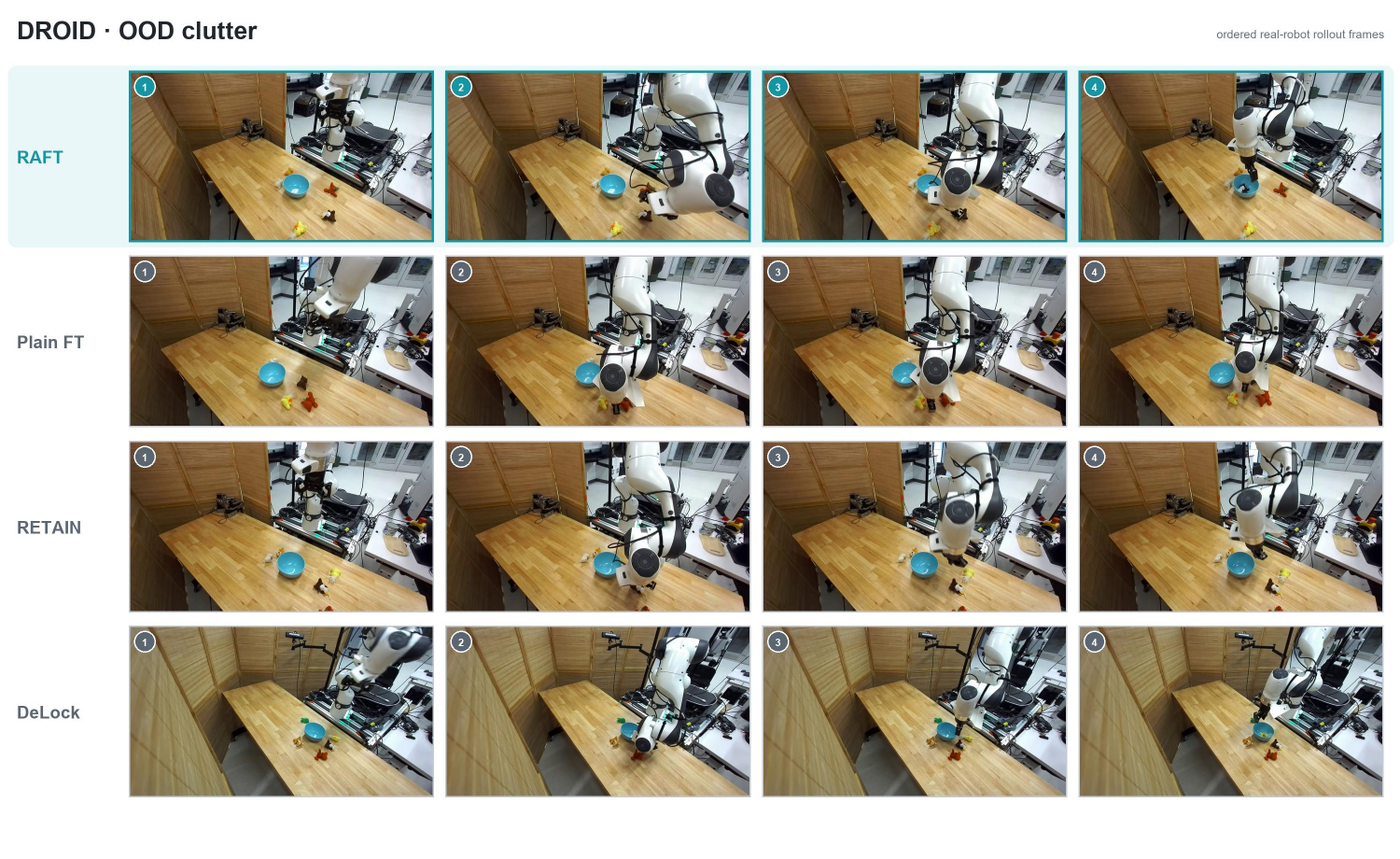}
\caption{\textbf{DROID rollout comparison under OOD clutter.} Rows show \ours{}, Plain FT, RETAIN, and DeLock at four ordered points in matched real-robot episodes. \ours{} completes the manipulation, while the comparison policies fail before task completion.}
\label{fig:droid-rollout-clutter}
\end{figure*}

\begin{figure*}[p]
\centering
\includegraphics[width=\textwidth]{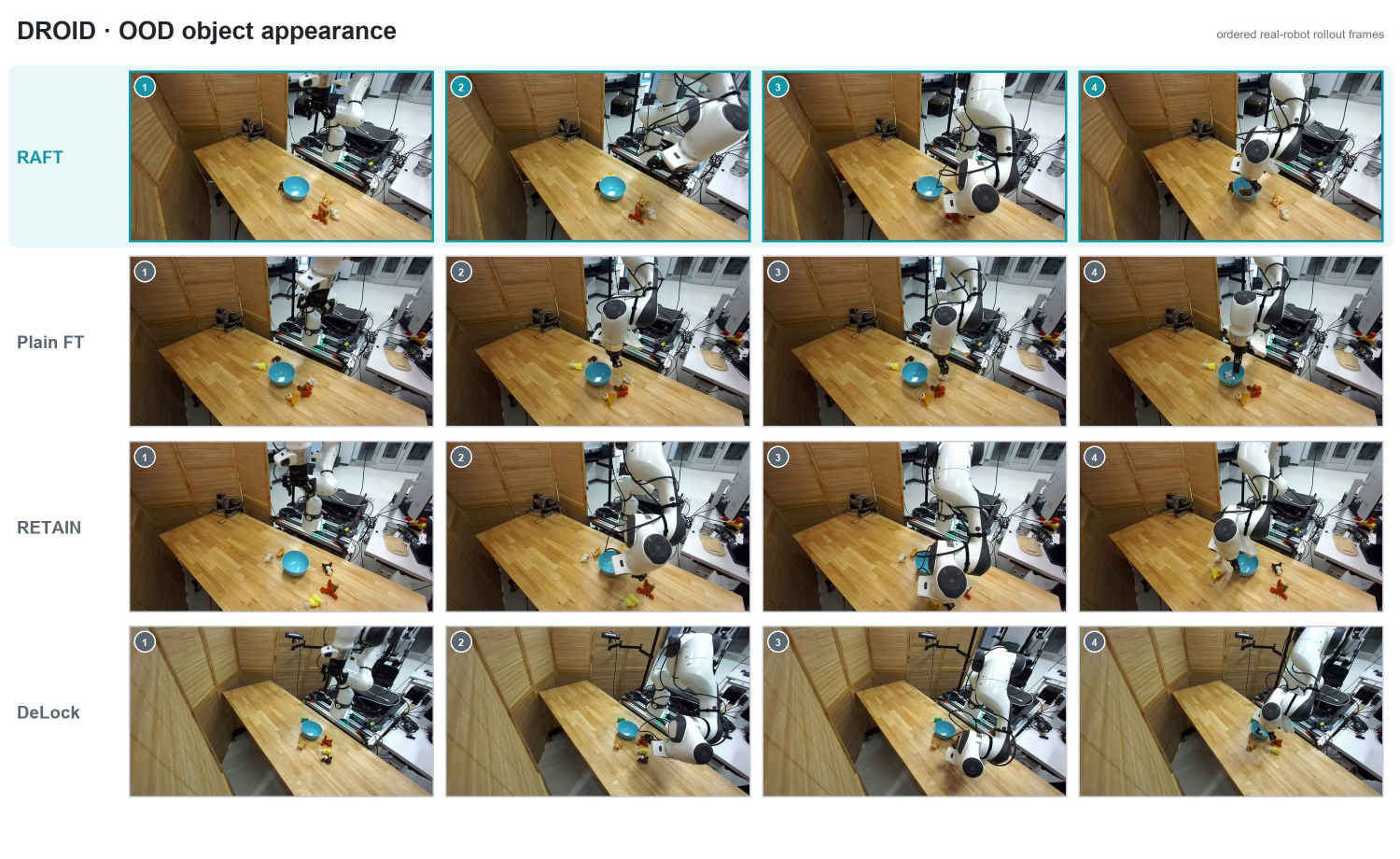}
\caption{\textbf{DROID rollout comparison with an OOD object.} Rows show \ours{}, Plain FT, RETAIN, and DeLock at four ordered points in matched real-robot episodes. \ours{} completes the manipulation, while the comparison policies fail before task completion.}
\label{fig:droid-rollout-object}
\end{figure*}

Figures~\ref{fig:droid-rollout-background}--\ref{fig:droid-rollout-object} compare policies on matched DROID evaluation cases in which \ours{} completes the task and the comparison policies fail before completion. Each method is shown at four ordered points in its rollout. Episode names and initial object configurations are matched across methods.

\begin{figure*}[p]
\centering
\includegraphics[width=\textwidth]{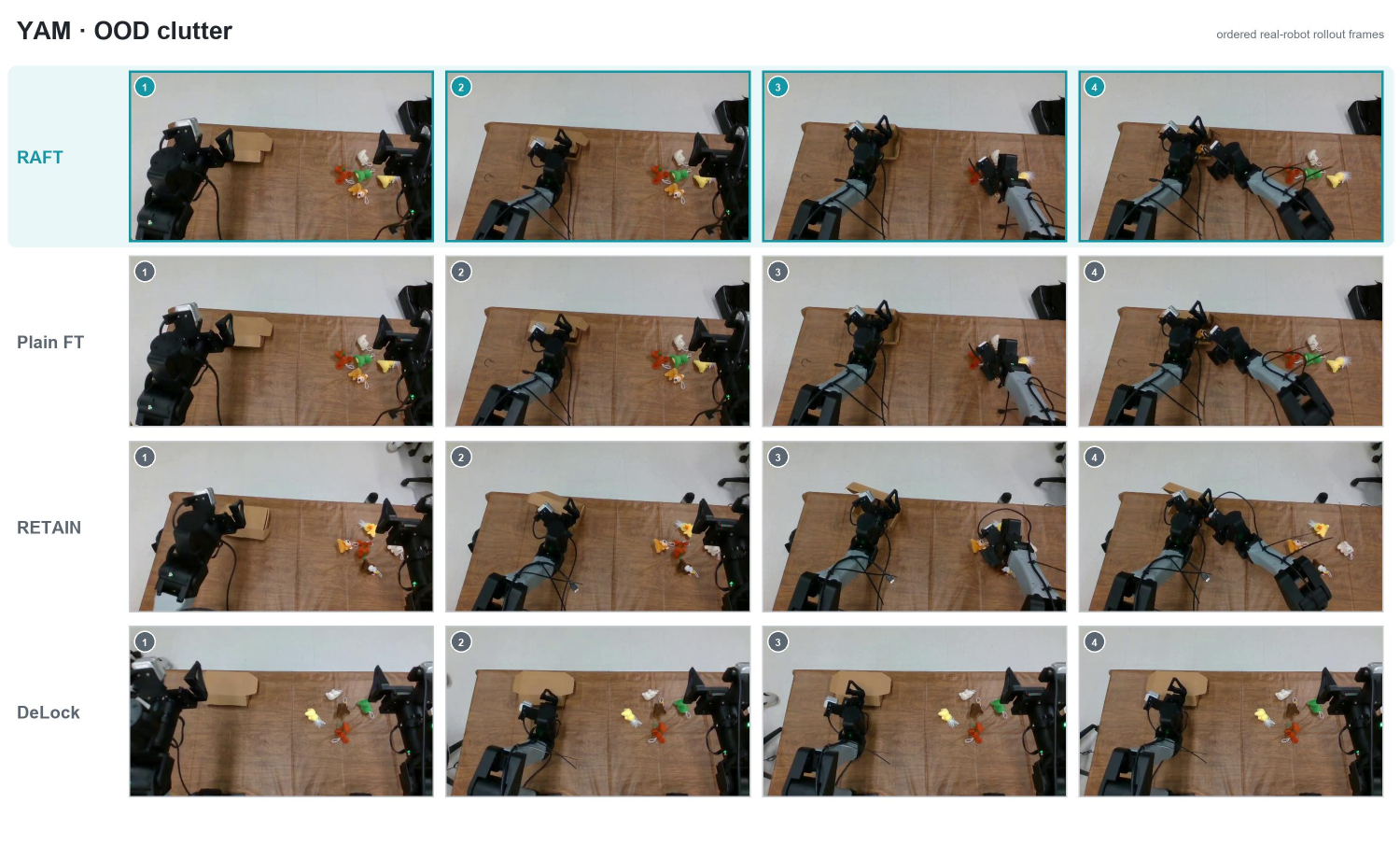}
\caption{\textbf{YAM rollout comparison under OOD clutter.} Rows show \ours{}, Plain FT, RETAIN, and DeLock on matched task instances; columns show ordered rollout frames. \ours{} completes the two-stage task, while the comparison policies do not.}
\label{fig:yam-rollout-clutter}
\end{figure*}

\begin{figure*}[p]
\centering
\includegraphics[width=\textwidth]{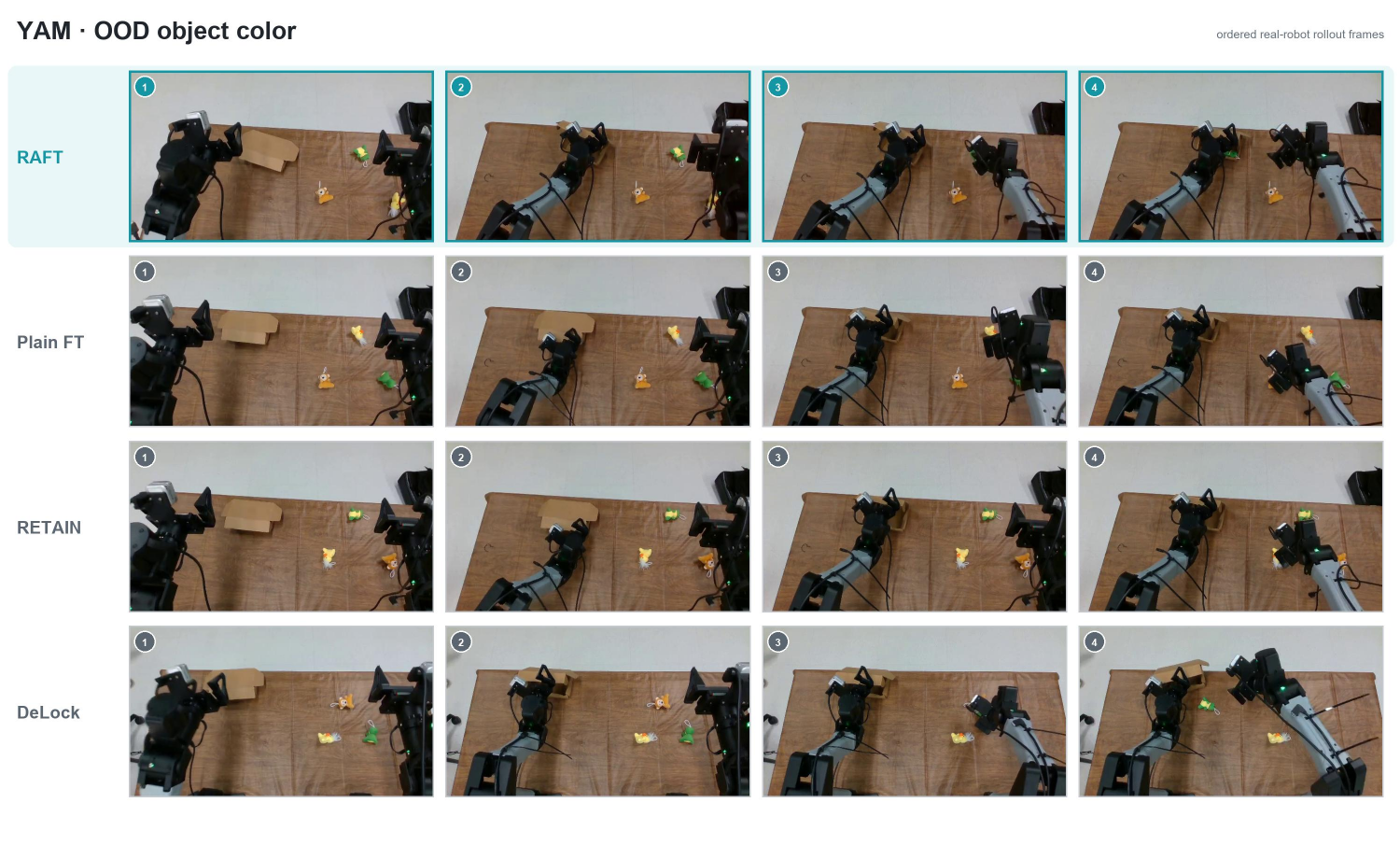}
\caption{\textbf{YAM rollout comparison under an OOD color shift.} Rows show \ours{}, Plain FT, RETAIN, and DeLock on matched task instances; columns show ordered rollout frames. \ours{} completes the two-stage task, while the comparison policies do not.}
\label{fig:yam-rollout-color}
\end{figure*}

\begin{figure*}[p]
\centering
\includegraphics[width=\textwidth]{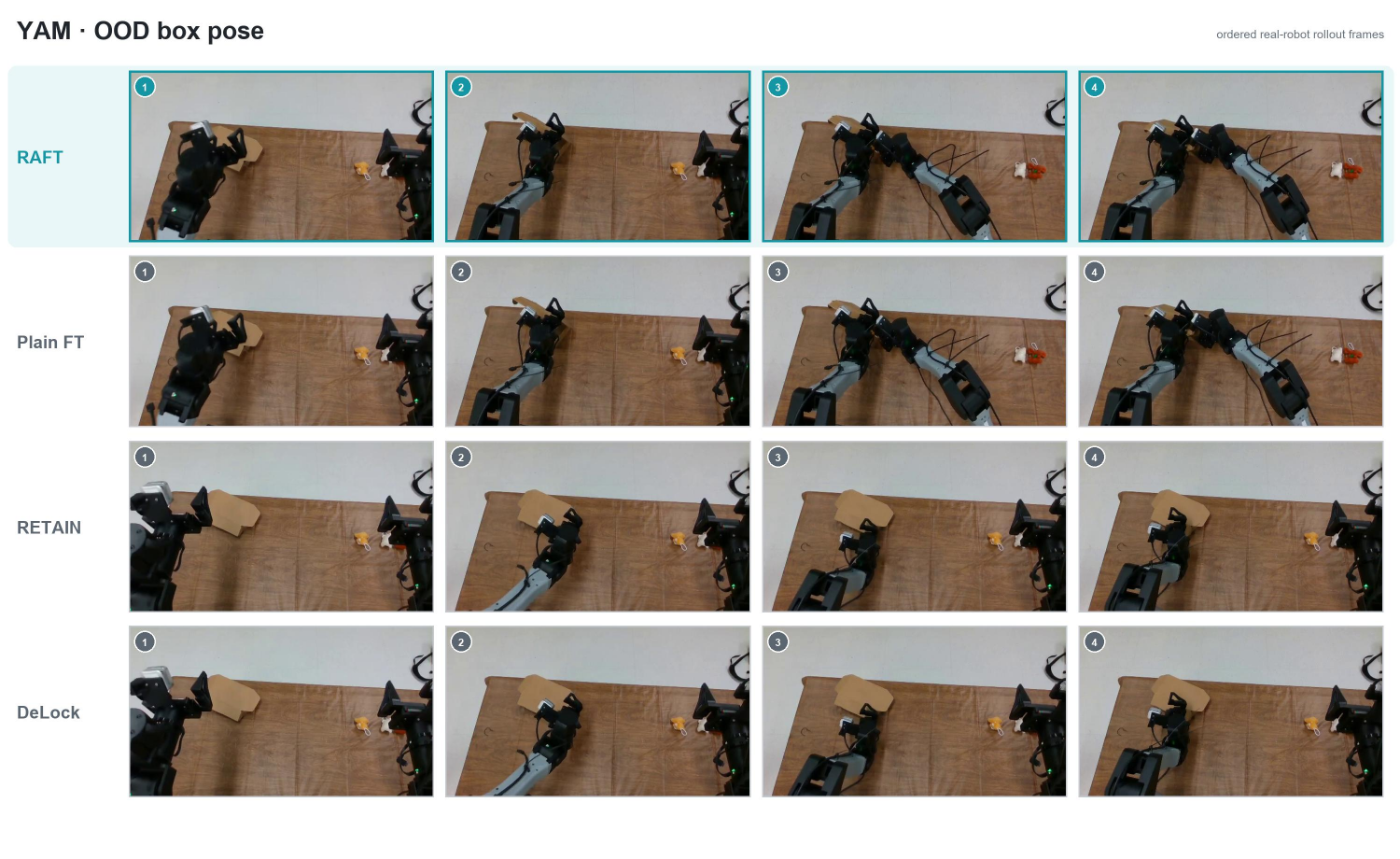}
\caption{\textbf{YAM rollout comparison under an OOD initial pose.} Rows show \ours{}, Plain FT, RETAIN, and DeLock on matched task instances; columns show ordered rollout frames. \ours{} completes the two-stage task, while the comparison policies do not.}
\label{fig:yam-rollout-pose}
\end{figure*}

Figures~\ref{fig:yam-rollout-clutter}--\ref{fig:yam-rollout-pose} show matched YAM evaluations under three OOD shifts. Rows correspond to \ours{}, Plain FT, RETAIN, and DeLock; columns follow each rollout in temporal order.

\section{Additional ablations and controls}
\label{app:ablations}

Section~\ref{subsec:modality} varies the format of the Stage-I supervision target; Section~\ref{sec:stage-ablation} ablates how the Stage-I reasoning adapter is treated during Stage II; Section~\ref{sec:cotrain-control} compares staged training with co-training and with frozen-vision fine-tuning; and Section~\ref{sec:retain-raft-control} tests whether \ours{} composes with RETAIN weight merging.

\subsection{Effect of the reasoning modality}
\label{subsec:modality}

\begin{figure*}[!tb]
\centering
\includegraphics[width=0.32\textwidth]{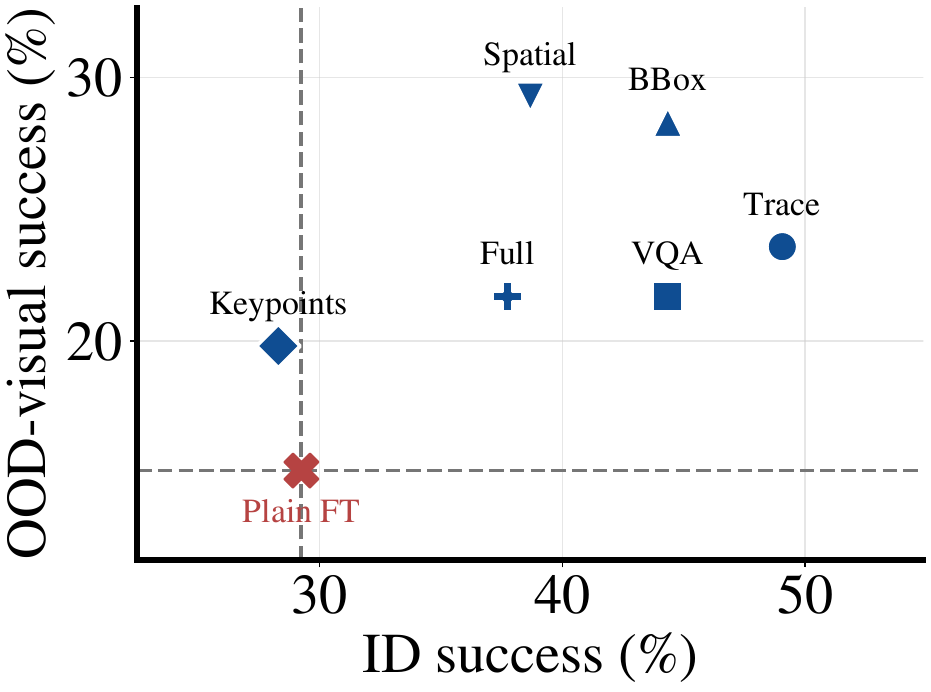}\hfill
\includegraphics[width=0.32\textwidth]{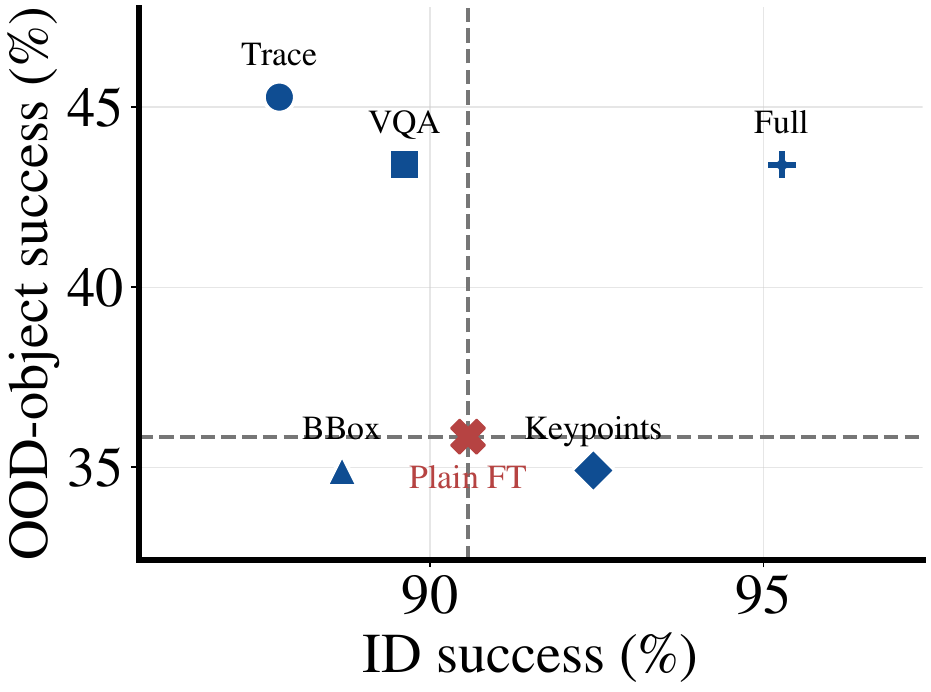}\hfill
\includegraphics[width=0.32\textwidth]{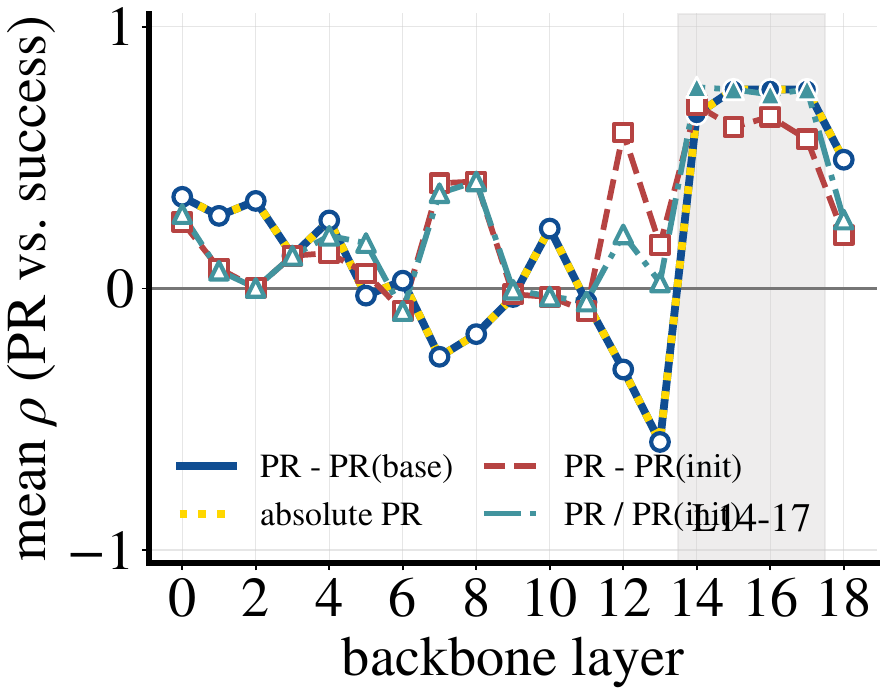}
\caption{\textbf{Effect of reasoning modality on success and representation dimensionality.} \emph{Left:} ID and OOD-visual success on the open drawer task for six reasoning modalities. \emph{Center:} ID and OOD-object success on pick and place for five modalities. Red crosses denote no-reasoning controls, and dashed lines mark Plain FT. Each policy is evaluated on 106 trials per axis. \emph{Right:} Layer-wise mean Spearman correlation between participation ratio and success, pooled over three task-matched evaluation axes and six reasoning variants, for four participation-ratio measures. Filled markers denote significance after FDR correction; shading marks layers 14--17.}
\label{fig:modality}
\label{fig:pr_analysis}
\end{figure*}

\paragraph{Setup.} We vary the Stage-I supervision target across six formats: full reasoning, visual question answering (VQA), spatial descriptions, bounding boxes, keypoints, and trajectory traces; Table~\ref{tab:stage1-modality-examples} gives an example of each. Each variant changes only the format of the Stage-I target, while Stage-II action training and the evaluation protocol are held fixed. The main-text results use full reasoning for pick and place and bounding-box supervision for the open drawer task, with the modality fixed across evaluation perturbations within each task; the remaining modalities are reported here. Figure~\ref{fig:modality} summarizes success on both tasks together with a layer-wise participation-ratio analysis.

\paragraph{Results.} On the \emph{open drawer} task, trajectory-trace supervision achieves the highest ID success ($49.06\%$ vs.\ $29.25\%$ for Plain FT). Under the visual shift, spatial descriptions, bounding boxes, and trajectory traces reach $29.25\%$, $28.30\%$, and $23.58\%$, respectively. On \emph{pick and place}, full reasoning achieves the highest ID success ($95.28\%$), whereas trajectory traces achieve the highest OOD-object success ($45.28\%$), followed by VQA and full reasoning ($43.40\%$ each). Bounding-box and keypoint supervision each reach $34.91\%$, comparable to Plain FT ($35.85\%$).

\subsection{Trainable reasoning adapter in Stage II}
\label{sec:stage-ablation}

\paragraph{Setup.} We evaluate the same Stage-I reasoning LoRA under three settings: (i) attached to the base policy without Stage-II action fine-tuning, (ii) updated during Stage II, and (iii) frozen during Stage II, as in \ours{}. Table~\ref{tab:stage-ablation} reports pick-and-place success over 106 episodes per split.

\begin{table*}[t]
\centering
\caption{\textbf{Stage-II reasoning-adapter ablation on pick and place.} Success rate (\%) when the Stage-I reasoning LoRA is applied without Stage II, updated during Stage II, or frozen during Stage II (\ours{}). Each split contains 106 episodes.}
\label{tab:stage-ablation}
\scriptsize
\setlength{\tabcolsep}{5pt}
\begin{tabular}{lcccccc}
\toprule
Adapter setting & ID & OOD-light & OOD-texture & OOD-object & OOD-prompt & Mean \\
\midrule
LoRA on base (no Stage II) & 72.64 & 60.38 & 9.43 & 33.02 & 66.04 & 48.30 \\
LoRA trainable in Stage II & 90.57 & 86.79 & 65.09 & 34.91 & \textbf{92.45} & 73.96 \\
LoRA frozen in Stage II (\ours{}) & \textbf{95.28} & \textbf{93.40} & \textbf{92.45} & \textbf{43.40} & 91.51 & \textbf{83.21} \\
\bottomrule
\end{tabular}
\end{table*}

\paragraph{Results.} Attaching the LoRA to the base policy without Stage II yields $48.30\%$ mean success. Updating the LoRA during Stage II raises mean success to $73.96\%$, and freezing it raises it further to $83.21\%$. The gap between the trainable and frozen settings is largest under the texture shift ($65.09\%$ vs.\ $92.45\%$).

\subsection{Comparison with co-training and frozen-vision fine-tuning}
\label{sec:cotrain-control}

\paragraph{Setup.} We compare staged training with co-training, in which reasoning and action supervision are applied within a single training run. Reasoning and action examples are sampled in a 50/50 ratio, and their losses are weighted equally. The primary Plain FT and co-training baselines fine-tune the SigLIP vision encoder. We also evaluate Plain FT with a frozen vision encoder, matching \ours{}, which adapts vision only through the Stage-I LoRA and keeps that LoRA frozen in Stage II. Table~\ref{tab:cotrain-control} reports pick-and-place success on the same 106 evaluation episodes per split.

\begin{table*}[t]
\centering
\caption{\textbf{Co-training and frozen-vision controls on simulated pick and place.} Success rate (\%) on five matched evaluation splits ($n=106$ per cell).}
\label{tab:cotrain-control}
\small
\setlength{\tabcolsep}{5pt}
\begin{tabular}{lcccccc}
\toprule
Method & ID & OOD-light & OOD-texture & OOD-object & OOD-prompt & Mean \\
\midrule
Plain FT & 90.57 & 87.74 & \textbf{92.45} & 35.85 & \textbf{91.51} & 79.62 \\
Plain FT (frozen vision) & 91.51 & 86.79 & 87.74 & 40.57 & \textbf{91.51} & 79.62 \\
Co-train & 79.25 & 70.75 & 72.64 & 17.92 & 75.47 & 63.21 \\
\ours{} & \textbf{95.28} & \textbf{93.40} & \textbf{92.45} & \textbf{43.40} & \textbf{91.51} & \textbf{83.21} \\
\bottomrule
\end{tabular}
\end{table*}

\paragraph{Results.} Co-training reaches $63.21\%$ mean success, compared with $79.62\%$ for Plain FT and $83.21\%$ for \ours{}; its largest deficit is under the object shift ($17.92\%$ vs.\ $43.40\%$ for \ours{}). Freezing the vision encoder in Plain FT changes its split-level results but leaves its mean success unchanged at $79.62\%$. Its OOD-object success is $40.57\%$, compared with $35.85\%$ for the primary Plain FT and $43.40\%$ for \ours{}.

\subsection{Compatibility with weight merging}
\label{sec:retain-raft-control}

\paragraph{Setup.} We test whether \ours{} can serve as the adapted checkpoint in RETAIN weight merging. Both merged policies interpolate with the same pretrained base at $\alpha=0.9$: one uses Plain FT as the adapted endpoint, and the other uses \ours{} with bounding-box supervision. We also evaluate the standalone \ours{} policy. Table~\ref{tab:retain-raft-open} reports results on the Franka open drawer task; every cell contains 106 trials.

\begin{table*}[t]
\centering
\caption{\textbf{RETAIN applied to Plain FT and \ours{} on the Franka open drawer task.} Success rate (\%) for each axis. RETAIN applied to \ours{} outperforms RETAIN applied to Plain FT.}
\label{tab:retain-raft-open}
\small
\setlength{\tabcolsep}{5pt}
\begin{tabular}{lcccc}
\toprule
Policy & ID & OOD-light & OOD-texture & Overall \\
\midrule
RETAIN on Plain FT ($\alpha=0.9$) & 38.68 & 34.91 & 17.92 & 30.50 \\
RETAIN on \ours{} (bbox, $\alpha=0.9$) & 41.51 & 37.74 & \textbf{29.25} & 36.16 \\
\ours{} (bbox) alone & \textbf{44.34} & \textbf{41.51} & 28.30 & \textbf{38.05} \\
\bottomrule
\end{tabular}
\end{table*}

\paragraph{Results.} Using \ours{} instead of Plain FT as the adapted endpoint increases pooled success from $30.50\%$ to $36.16\%$. Standalone \ours{} reaches $38.05\%$, the highest overall, with the highest ID and lighting success ($44.34\%$ and $41.51\%$).

\setlength{\tabcolsep}{4pt}

\section{Evaluation environments}
\label{app:evaluation-conditions}

\begin{figure*}[t]
\centering
\includegraphics[width=\textwidth]{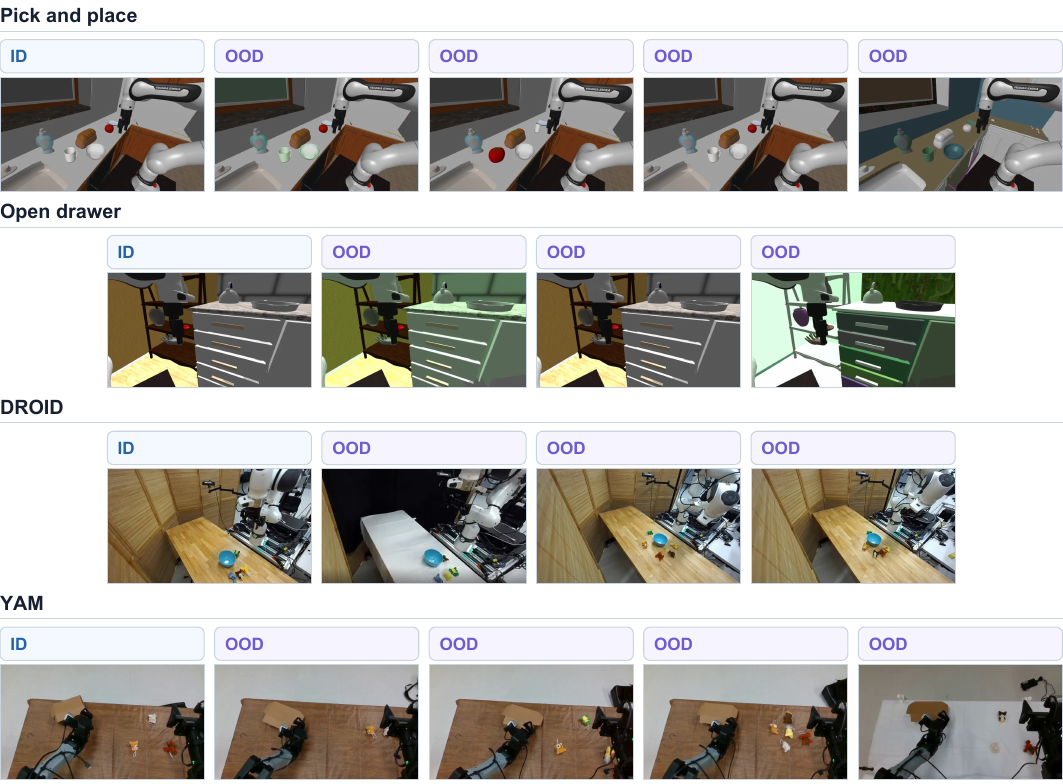}
\caption{\textbf{Evaluation setup across the four robot tasks.} Each row begins with an in-distribution observation, followed by its OOD variants. Pick and place varies lighting, target object, instruction, and scene texture; the open drawer task varies lighting, instruction, and scene texture; DROID varies background, clutter, and object appearance; and YAM varies box pose, object color, clutter, and background. Instruction shifts reuse the ID image because only the task prompt changes.}
\label{fig:supp-evaluation-conditions}
\end{figure*}

Figure~\ref{fig:supp-evaluation-conditions} illustrates the in-distribution (ID) and out-of-distribution (OOD) conditions used in our simulated and physical-robot experiments; the evaluation protocol is described in Appendix~\ref{app:evaluation-protocol}. Simulation task construction, resets, initial-state matching, and success criteria follow MolmoSpaces~\cite{kim2026molmospaces}. Within each perturbation, all methods are evaluated on identical episodes. On the physical robots, resets preserve the task configuration and initial state.

\section{Representation and attention analysis protocol}
\label{app:diagnostics-protocol}

\paragraph{Model and captured frames.} The participation-ratio (PR) analysis in Figure~\ref{fig:pr_analysis} uses the first 400 records of the held-out iTHOR FloorPlan13 validation split, spanning five episodes. The linear-probe results in Figure~\ref{fig:late_layer_probes} use a  capture of 414 frames: nine evenly spaced frames from each of 46 policy-training episodes. 

\paragraph{Participation ratio and correlations.} For each frame and decoder layer $\ell$, we mean-pool \texttt{hidden\_states[$\ell$]} over non-padding positions to obtain a $400\times2048$ matrix $X_\ell$. After centering the columns, we compute the covariance eigenvalues $\lambda_i$ from the squared singular values and define $\mathrm{PR}_\ell=(\sum_i\lambda_i)^2/\sum_i\lambda_i^2$. PR thus measures the effective dimensionality of \emph{across-frame} variance. Each evaluation axis pairs one downstream benchmark with one PR probe domain. Figure~\ref{fig:pr_analysis} pools the three task-matched axes. For each layer and axis, we compute the Spearman correlation across the same six Stage-I arms and assess significance with a two-sided exact permutation test over all $6!=720$ assignments, applying Benjamini--Hochberg FDR correction across the 19 layers.

\paragraph{Linear probes.} Episode-time labels are defined as $\min(3,\lfloor4\,\mathrm{frame\_idx}/\mathrm{total\_frames}\rfloor)$, giving a chance balanced accuracy of $1/4$. The labels \emph{reach}, \emph{grasp}, \emph{transport}, and \emph{place} denote the four positional (timestep) quartiles of each episode. Within each leave-one-episode-out fold, we fit \texttt{StandardScaler}, \texttt{PCA(64)}, and \texttt{LogisticRegression(C=0.5)}. Layer-band uncertainty and differences from Plain FT are estimated with a paired bootstrap over episodes ($B=1000$).

\paragraph{Visual attention and region masks.} At each layer, we extract the attention from text queries to image keys, $A[:,256:,:256]$, average it uniformly over heads and text queries, and $L_1$-normalize it over the 256 image patches; no individual head or last-token readout is selected. Band-level maps average the normalized per-layer maps. Because this normalization removes the total attention mass on the image, we report that quantity separately as \texttt{img\_mass}. For visualization, patch values are upsampled by nearest-neighbor tiling, and all arms share a color scale capped at the 99th percentile of interior-patch attention; the border ring is excluded from this scale because it exhibits attention-sink/register behavior. Bounding boxes are mapped to patches using floor/ceiling boundaries and may include up to one extra patch on each edge; each keypoint or trace point is mapped to the patch that contains it, and the background is the complement of all annotated patches. A frame enters a given region analysis only if it contains at least two bounding-box patches, one keypoint, or one trace point, as applicable. For a patch mask $M$ and normalized attention $p_j$, we define enrichment as $(256/|M|)\sum_{j\in M}p_j$, so that a value of 1 corresponds to uniform attention. Bounding-box, keypoint, and trace masks may overlap and therefore do not partition the image; only the background mask is defined by complementation.

\section{Extended late-layer probe analysis}
\label{app:all_arm_probes}
\begin{figure}[ht]
  \centering
  \includegraphics[width=\linewidth]{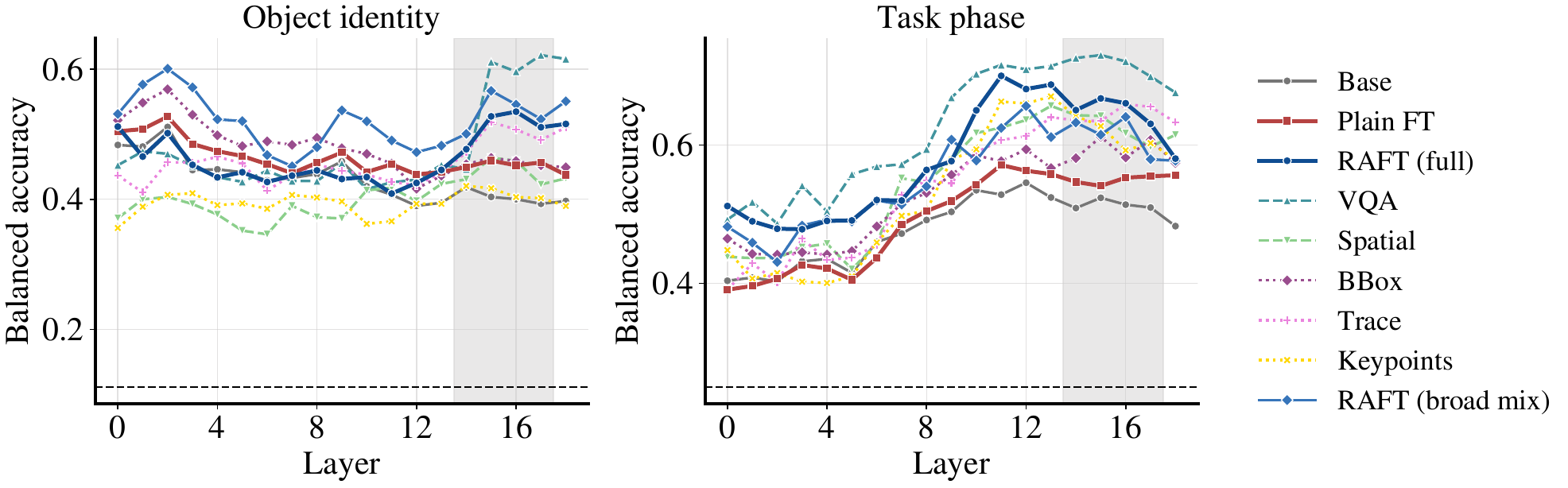}
  \caption{\textbf{Late-layer probes for all nine model arms.} Leave-one-episode-out balanced accuracy for object identity and episode-time quartile across backbone layers. The shaded band marks layers 14--17. This expands the four-arm comparison in the main text to the full set of reasoning-modality variants.}
  \label{fig:probe_all_arms}
\end{figure}

Figure~\ref{fig:probe_all_arms} expands the main-text probe comparison to all nine model arms. VQA and trace show the largest late-layer recovery in object-identity decoding and the strongest novel-object results; spatial, bounding-box, and keypoint variants show no comparable recovery and are the weakest on that shift.

\end{document}